\pdfoutput=1
\documentclass{article}

\PassOptionsToPackage{table}{xcolor}
\usepackage[authoryear,sort&compress,round]{natbib}
\usepackage[preprint,logo]{agibot_enerverse}
\usepackage{booktabs,tabularx,array}
\usepackage{booktabs,tabularx,array,xcolor,colortbl,multirow}

\usepackage[utf8]{inputenc}
\usepackage[T1]{fontenc}
\usepackage{amsmath,amsfonts,amssymb}
\usepackage{booktabs,multirow,longtable}
\usepackage{caption,subcaption,wrapfig,float,placeins,afterpage}
\usepackage{algorithm,algpseudocode}
\usepackage{microtype,xspace,fontawesome5}

\providecommand{\footerfont}{\normalfont}

\title{Learning to Live: A Step Toward Lifelong Embodied Intelligence}
\author{%
AgiBot Yuanzheng Research Team\\[0.35em]
{\small\normalfont\href{https://github.com/AgibotTech/LIFE}{\faGithub\enspace AgibotTech/LIFE}}\\[0.2em]
{\small\normalfont\href{https://agibotgeneral.github.io/life-site/}{\faGlobe\enspace Project Website}}}
\date{October 2026}
\hypersetup{pdfauthor={AgiBot Yuanzheng Research Team},pdftitle={Learning to Live: A Step Toward Lifelong Embodied Intelligence}}

\begin{document}
\maketitle
\vspace{-14pt}

\begin{abstract}
Embodied agents are usually organized around individual tasks, yet the physical world keeps changing between tasks and during execution. We introduce \textbf{LIFE (Lifelong Intelligence from Experience)}, a framework in which a robot observes continuously, remembers selectively, and turns experience into new capabilities. \textbf{RoboBeats}, a family of lightweight streaming gates, tracks conditions set by the agent and discovers new events, deciding when an observation warrants a response or a memory write. Selected observations enter \textbf{Plastic Memory}, a flat store of images and structured records that keeps the source evidence, so new fields and indexes can be added when later questions need them. When existing tools fall short, \textbf{Explore--Distill--Mount (EDM)} turns solutions found during problem solving into validated, reusable tools and skills. Under matched protocols, RoboBeats substantially improves condition tracking over its base models and, at the same writing budget, retains more useful evidence for memory question answering than uniform sampling. A case study of limited scale, replaying 24 hours of continuous video recorded by a real robot, illustrates how the integrated system builds memory from a day of experience and uses it to answer everyday questions. LIFE takes a step toward robots that become more familiar with their surroundings and more capable through experience: learning not only to complete tasks, but to live in the world.
\end{abstract}

\begin{figure}[h]
    \centering
    \includegraphics[width=\linewidth]{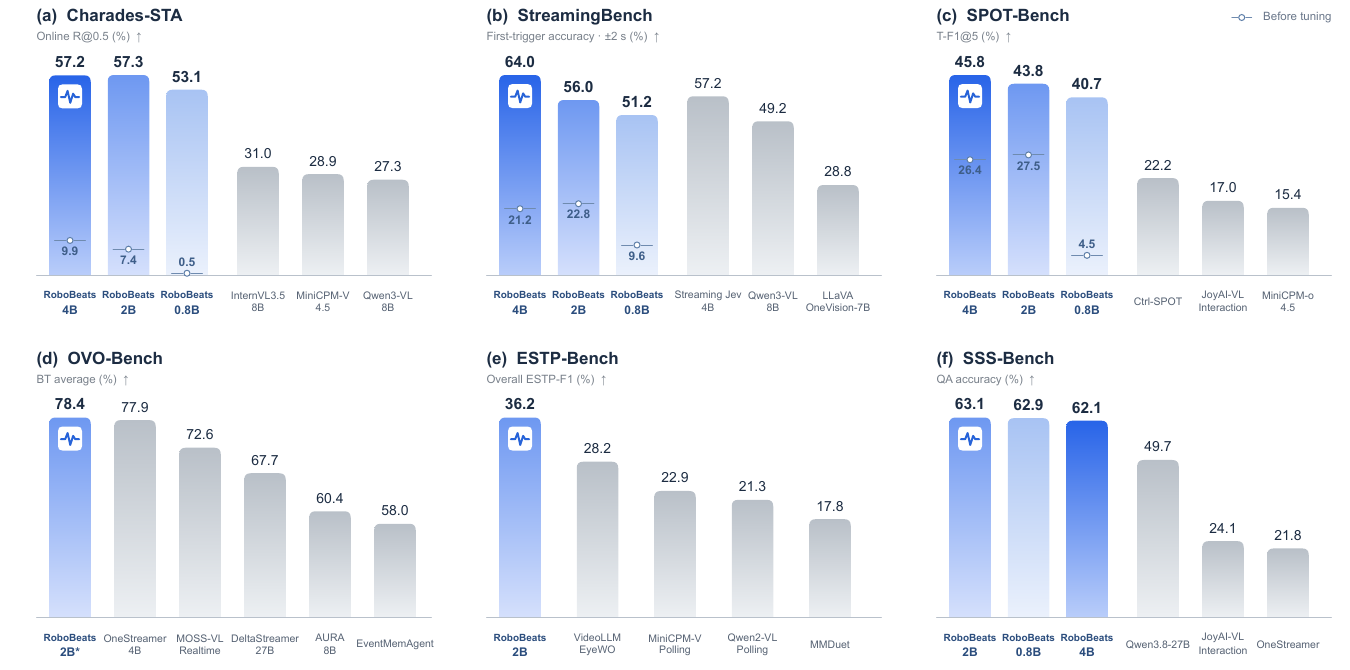}
    \caption{\textbf{RoboBeats across streaming perception and memory question answering benchmarks.}
    Results cover temporal localization (Charades-STA), condition triggering
    (StreamingBench), action-boundary responses (SPOT-Bench), proactive
    response (ESTP-Bench), and retrospective question answering
    (OVO-Bench and SSS-Bench).  Each panel uses its benchmark-specific metric; higher is better. RoboBeats-2B* uses a OneStreamer-4B reader. Details are given in Section~\ref{sec:robobeats-evaluation}.}
    \label{fig:robobeats-evaluation}
\end{figure}

\section{Introduction: From Individual Tasks to Lifelong Embodied Agents}

We hope for robots to participate in everyday human environments over extended periods, understanding changes, accumulating experience, and gradually learning to handle new situations. A robot on its hundredth day in a building should be more familiar with it than on its first: the people it has encountered, the places it has visited, and the problems it has handled should help it make better judgments and find more effective solutions. We call this goal \emph{Learning to Live} and study how embodied agents can continually build knowledge and capabilities through interaction with the world.

Many embodied-agent systems are organized around user-specified tasks: they interpret an instruction, select or compose skills, and execute actions toward the requested goal~\citep{ahn2022saycan,huang2022innermonologue}. Yet the physical world continues to change both between tasks and during execution. Objects may move, people may arrive, and seemingly incidental events may become relevant later. This creates a mismatch between the agent's discrete reasoning and tool calls and the world's continuous evolution. Scene descriptions, success detection, and human feedback can help agents revise their plans during execution~\citep{huang2022innermonologue}. A relevant change may require a response before the next planned decision, while a completed tool call may leave the intended physical outcome uncertain: reaching a location, for example, does not confirm that the object sought is still there. The agent therefore needs ongoing observations to update its understanding, bring relevant changes into its decision process, and verify outcomes throughout interaction.

Sustained observation must also be practical over long periods. Continuous visual streams contain both brief, consequential events and large amounts of repetitive information, while the relevance of an observation to future tasks may not yet be known~\citep{zeng2026onestreamer,resight_worth}. Processing every observation with an expensive model would increase computational cost and response latency, while retaining and indexing every observation without selection would make memory increasingly redundant. The system must identify when further analysis is warranted and decide what experience to preserve. At the same time, memory should retain enough evidence to support future questions that were not anticipated when an observation was made.

Recent work has explored continuous visual interaction and proactive memory, including systems that decide when to respond or record evidence from video streams~\citep{yao2026joyai,zeng2026onestreamer}. Embodied agents have also used long-term visual and spatial memory to answer questions and guide navigation~\citep{resight_remembr,resight_raven}, while systems such as BetaMem and ABot-AgentOS study memory evolution and integrated robot reasoning~\citep{liu2026betamem,tian2026abot}. These approaches provide a basis for studying how ongoing perception, experience, and capability development can work together in a robot system.

To address these challenges, we introduce \textbf{LIFE (Lifelong Intelligence from Experience)}, a framework that connects proactive perception, long-term memory, reasoning, and autonomous learning. LIFE continuously screens observations for changes relevant to the current context or potentially useful in the future. It turns selected experiences into retrievable historical evidence and uses that evidence to support decisions, verification, and further exploration. As the agent encounters problems that existing methods cannot handle, it can also develop and retain new tools and skills, allowing accumulated experience to contribute to future capabilities.

LIFE's proactive perception is implemented by \textbf{RoboBeats}, a lightweight gate that tracks specified conditions and detects new events in streaming video. It decides when an observation should trigger a response, further perception, or memory writing, so that a large model does not have to analyze every frame. Figure~\ref{fig:robobeats-evaluation} summarizes its evaluation across streaming perception and memory question answering benchmarks.

Selected experiences enter \textbf{Plastic Memory}, which stores observations with their temporal and spatial context and keeps the source images~\citep{resight_remembr,resight_dynamem}. Structured records give fast access through people, places, objects, and events, while the retained images let the agent recover details that were not extracted when the experience was recorded. Writing only selected observations keeps memory compact, and keeping the images leaves it usable for questions that were not anticipated.

The information needed for a new problem may already exist in retained observations even when the agent lacks a way to extract or connect it. Building on skill accumulation in embodied agents~\citep{resight_voyager,resight_aspire}, LIFE uses the \textbf{Explore--Distill--Mount (EDM)} mechanism from ReSight~\citep{che2026resight} to turn solutions developed for individual problems into validated tools and skills that apply to both past experience and future tasks.

This report presents the design of LIFE and its core components: RoboBeats for proactive perception, Plastic Memory for retaining and revisiting experience, and EDM for constructing and reusing tools and skills. We evaluate perception gating through streaming benchmarks, examine memory retrieval and capability reuse, and present a case study of limited scale that replays a day of real robot video to examine how the integrated system builds and uses memory. Together, these studies investigate how an embodied agent can observe a changing world, make use of its experience, and extend what it can do over time.

\section{LIFE Overview: From Experience to Knowledge and Capabilities}
\label{sec:life-overview}

\begin{figure}[t]
    \centering
    \includegraphics[width=\linewidth]{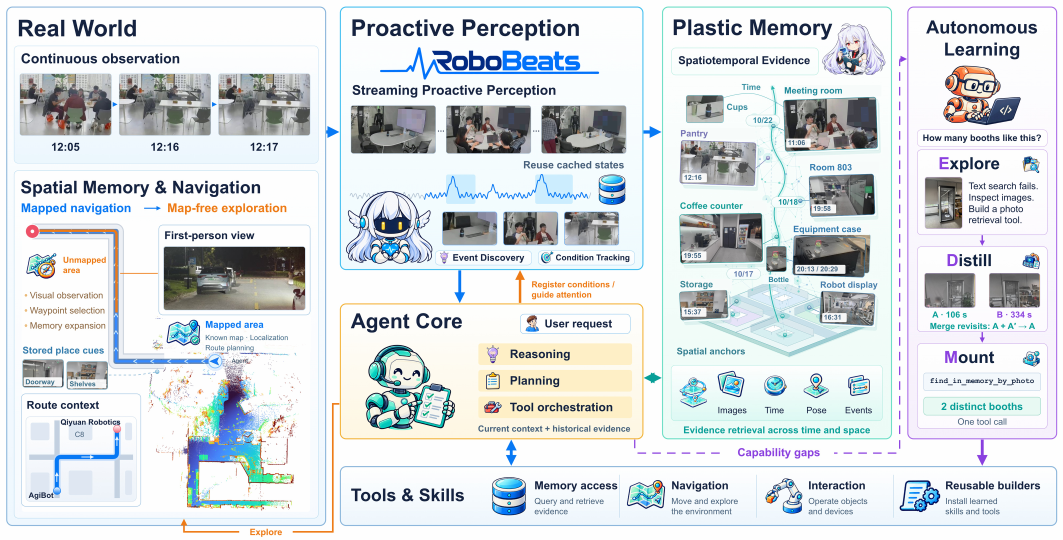}
    \caption{Overview of LIFE. Continuous perception detects changes in the environment, while active exploration extends observation coverage. Together, they accumulate memories with temporal and spatial information. The agent core combines the current context with historical evidence to reason, plan, and invoke tools, and in turn directs attention and exploration. Autonomous learning develops candidate methods to address capability gaps encountered through experience and adds validated methods to the tool and skill layer.}
    \label{fig:life-overview}
\end{figure}

LIFE connects continuous observation, persistent memory, and capability development through an agent that reasons and acts on its experience, building on work in proactive visual interaction and embodied memory~\citep{zeng2026onestreamer,resight_raven,liu2026betamem,tian2026abot}. As shown in Figure~\ref{fig:life-overview}, the agent sets attention conditions, queries Plastic Memory, and invokes tools to answer questions or act in the world. RoboBeats screens incoming observations for conditions that require a response and events worth recording. A writer turns selected observations into memory, which supplies evidence for subsequent reasoning. When the available tools cannot meet a new need, EDM provides a process for developing and retaining additional methods. The shared tool and skill layer exposes memory access, physical actions, and reusable methods to the agent.

\subsection{From Waiting for Instructions to Attending to the World}

RoboBeats keeps screening the visual stream during task execution and between user requests. Its output feeds two workflows: condition tracking supports responses to the current situation, and event discovery supplies observations for memory writing. Both use lightweight screening to decide when further processing is needed.

For condition tracking, the agent or a standing interaction rule specifies what to watch for, such as a person starting a phone call or the visible completion of a manipulation task. When the condition changes state, the system triggers a response or returns the observation to the agent for reassessment. Monitoring continues while an action tool is running, and the agent combines these observations with tool feedback and task context to decide whether to continue, finish, or revise its next action. Section~\ref{sec:robobeats-applications} shows this for an elevator door, a handover, a phone call, and a grasp.

For event discovery, RoboBeats identifies changes worth describing and passes the selected observations to the writer, which records each event in an \texttt{event} field with its supporting images and timestamps. A later consolidation step links related records about people, objects, and activities. Event discovery decides what enters memory; the writer decides how it is described. Section~\ref{sec:real-robot-memory-qa} applies this pipeline to a full day of robot video.

\subsection{Making an Unfamiliar World Familiar}
\label{sec:life-familiar}

Plastic Memory retains selected images and associated JSON records in a flat file structure. Each saved observation carries its acquisition time, its pose when available, and fields extracted from the evidence (Section~\ref{sec:life-memory-evidence}), which give access through people, places, objects, and events. The source images remain available for details the original descriptions did not capture.

Given a question, the agent retrieves candidate records, follows their temporal or semantic links, and inspects the supporting images when needed. When a question needs a detail that no field records, a program extracts it from the retained images and adds it to their records (Section~\ref{sec:life-memory-extension}). Because the memory is a set of ordinary files, programs and coding agents make these updates with standard file operations.

Navigation extends and refreshes this experience. A retrieved observation supplies a destination pose, the robot reaches it with map-based or vision-and-language navigation, and on-site perception checks the current state; frames collected along the way enter memory as new evidence.

\begin{figure}[t]
    \centering
    \includegraphics[width=\linewidth]{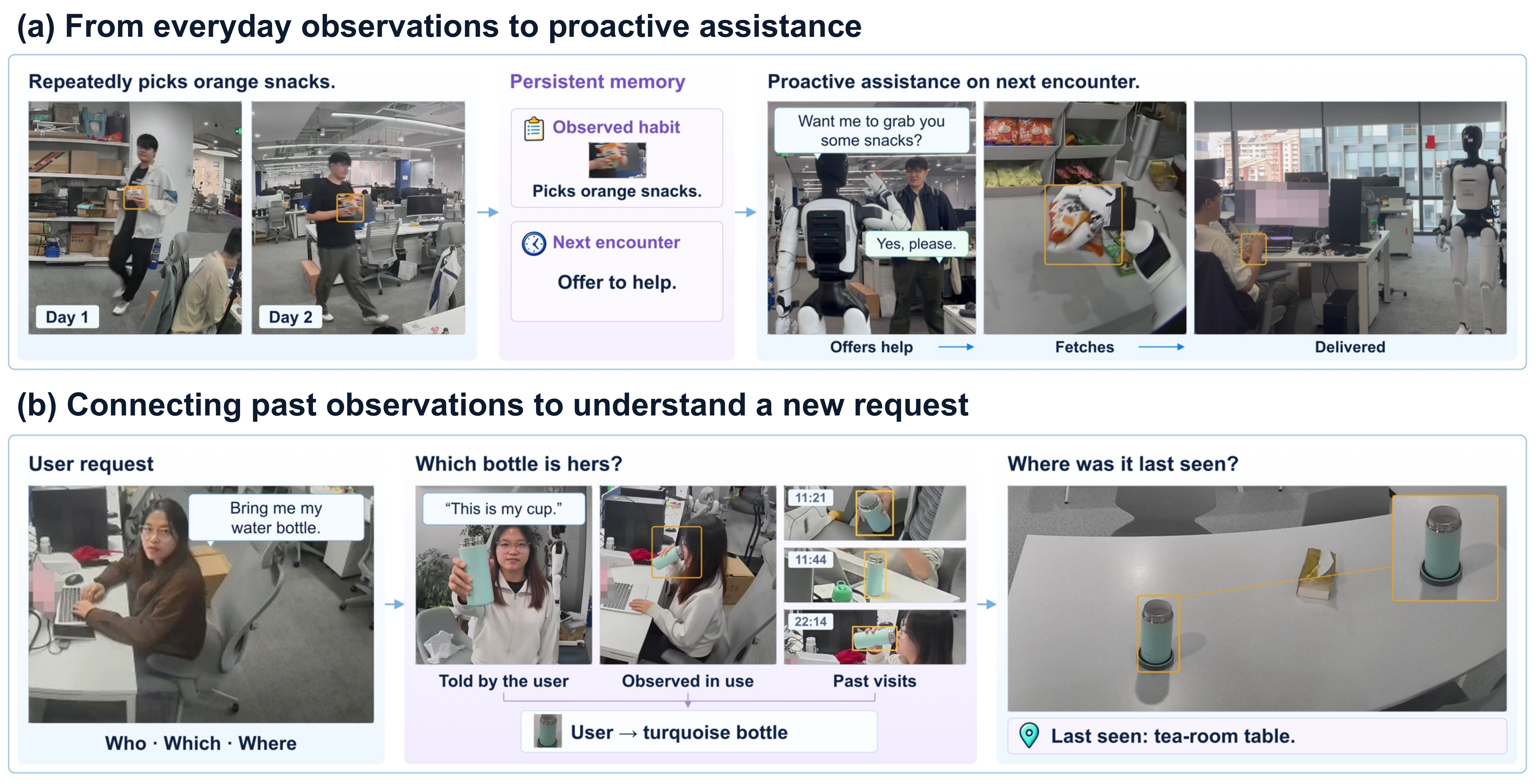}
    \caption{Everyday experience supports proactive and requested assistance. (a) Observations of snack-taking are retained and used to offer help on a later encounter, followed by fetching and delivery after the user agrees. (b) A request for ``my bottle'' is interpreted by connecting the user's statement, observed use, and past sightings to identify the bottle and retrieve its last recorded location.}
    \label{fig:life-experience-cases}
\end{figure}

Figure~\ref{fig:life-experience-cases} shows two uses of shared history: the robot notices a person's repeated snack-taking and offers help on a later encounter; it also uses earlier interactions and sightings to identify the user's bottle and where it was last seen.

\subsection{Turning Experience into Reusable Capabilities}

The agent first solves a problem with the evidence and tools already available: it retrieves structured fields, inspects source images, or gathers a new observation through physical action. Ways of extracting or using evidence that prove useful repeatedly become capabilities for later tasks.

EDM organizes this process as Explore--Distill--Mount. Exploration investigates a capability gap with historical evidence and candidate programs; distillation turns an effective method into a reusable tool or skill; mounting registers the validated method in the shared capability layer (Section~\ref{sec:life-edm}). A field-extraction method reprocesses earlier images and enriches Plastic Memory, while a retrieval method lets later tasks find evidence directly.

Let $\mathcal{M}_t$ denote persistent memory at time $t$ and $\mathcal{T}_t$ the available tools and skills. Experience updates both:
\begin{equation}
\begin{aligned}
    \mathcal{M}_{t+1} &= \operatorname{Update}(\mathcal{M}_t,\Delta\mathcal{E}_t),\\
    \mathcal{T}_{t+1} &= \mathcal{T}_t\cup\Delta\mathcal{T}_t^{\mathrm{validated}}.
\end{aligned}
\label{eq:life-update}
\end{equation}
Here, $\Delta\mathcal{E}_t$ contains new evidence and interpretations supported by it, while $\Delta\mathcal{T}_t^{\mathrm{validated}}$ contains validated methods retained for future use. Recording where a cup was last seen updates memory; developing a reusable way to find objects in historical observations expands the available capabilities.

The following sections develop these components and their use. Section~3 presents RoboBeats for condition tracking and selective memory writing. Section~4 introduces Plastic Memory and its access and extension mechanisms. Section~5 evaluates EDM's ability to develop and reuse tools and skills. Section~6 presents an integrated robot system using its established memory and tools to answer everyday questions about people, objects, and activities from a day of recorded experience.

\section{A Proactive Gate for Streaming Visual Perception}
\label{sec:active-perception}

\begin{figure}[t]
    \centering
    \includegraphics[width=\textwidth]{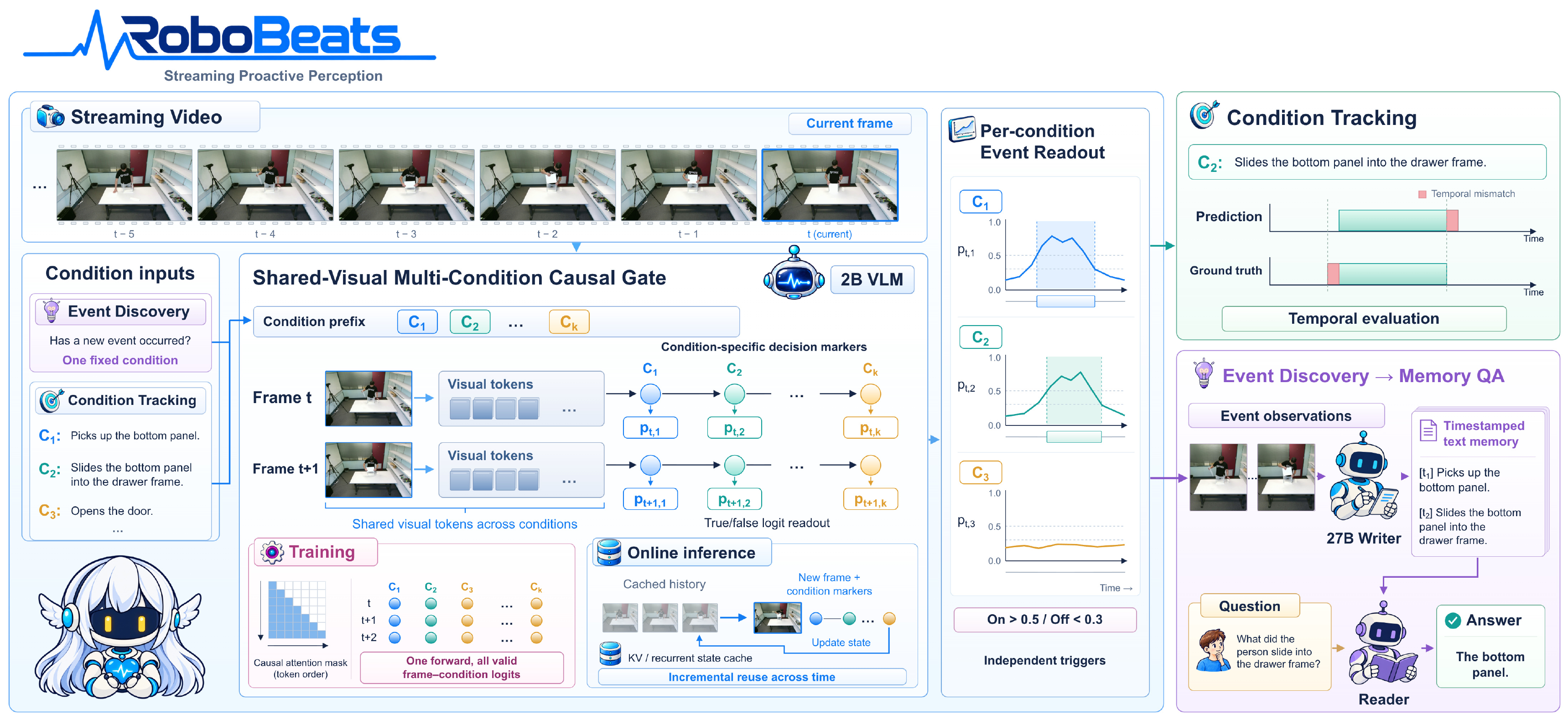}
    
\caption{\textbf{Overview of RoboBeats.} Conditions share a causal visual stream and receive independent predictions. \textbf{Condition tracking (top right)} compares predicted activation intervals with annotations. \textbf{Event discovery and memory question answering (bottom right)} connect selected observations to a writer and reader; future questions are not provided to the gate.}

\label{fig:overview}

\end{figure}

Robots in continuous operation need to observe their surroundings and determine which changes warrant further processing for the current task. A system may need to issue a timely reminder when a specified event occurs, or record newly observed activities before future tasks are known. Streaming video perception must balance the need for timely decisions with the computational cost of long-term operation.

Streaming systems have studied timely responses and query-independent memory formation~\citep{yao2026joyai,zeng2026onestreamer,wen2026eventmemagent}, while selective episodic memory has also been explored for robots~\citep{resight_worth}. ProactEgoMem evaluates proactive memory acquisition and later recall in egocentric video~\citep{liao2026proactegomem}.

We introduce \textbf{RoboBeats}, a family of lightweight decision models based on Qwen3.5 that serve as proactive gates for streaming visual perception. It processes frames in temporal order and evaluates whether a specified condition holds or an event has occurred. These judgments update its gating state, which determines when to pass observations to downstream modules. RoboBeats supports both condition tracking and event discovery, providing a basis for proactive responses and memory writing.

\subsection{Conditioned Streaming Perception}
\label{sec:robobeats-architecture}

RoboBeats formulates continuous perception as a sequence of judgments about natural language conditions. As frames arrive, the model scores each condition using the current frame and previous observations. It supports two modes. \textit{Condition tracking} monitors specific descriptions, such as ``someone opens a door,'' and judges whether each condition holds over time. \textit{Event discovery} uses the fixed query ``Has a new event occurred?'' to identify observations of new activities or changes. Both modes use the same input structure and scoring procedure. Event discovery is the case with a single monitored condition.

We develop 0.8B, 2B, and 4B variants using the Qwen3.5 family~\citep{qwen2026qwen35}. The different sizes allow a deployment to choose a model suited to the available on-robot compute and memory. The input sequence starts with the monitoring conditions, followed by timestamped images in temporal order. Each image is followed by a decision query for each condition, with its result read separately. Each image appears only once, so multiple conditions share its visual representation and observation history. The input is causal. Each decision uses the condition descriptions, the current image, and earlier observations within the bounded context. Predictions are maintained independently for each condition, allowing several conditions to hold simultaneously.

At the final position of each decision query, we read the logits for the single-token labels \texttt{true} and \texttt{false}. Let $I_t$ be the image at step $t$, with timestamp $\tau_t$, and let $\mathcal{C}=\{c_1,\ldots,c_K\}$ be the set of natural language conditions. For condition $c_k$, the two logits are $z^{\mathrm{true}}_{t,k}$ and $z^{\mathrm{false}}_{t,k}$. The normalized score is
\begin{equation}
p_{t,k} =
\frac{\exp\left(z^{\mathrm{true}}_{t,k}\right)}
{\exp\left(z^{\mathrm{false}}_{t,k}\right) +
 \exp\left(z^{\mathrm{true}}_{t,k}\right)}.
\label{eq:condition_score}
\end{equation}
A higher $p_{t,k}$ indicates stronger evidence that condition $c_k$ holds in the current observation. Reading the label logits avoids generating a textual answer for each frame, consistent with the broader use of direct structured judgments in System One models such as Jev, OmniJev, and VideoJev~\citep{almeida2026jev,omnijev2026,zhu2026videojev}. Condition tracking produces one score sequence per target condition. Event discovery produces a single sequence with $K=1$.

The gate converts condition scores into activation intervals. Interval onsets can trigger responses, while selected observations can be passed to downstream modules for memory writing (Figure~\ref{fig:overview}).

\subsection{Training and Online Inference}
\label{sec:robobeats-learning}

RoboBeats learns condition judgments from continuous frames and temporal labels. During online operation, it reuses previous computation to process new frames incrementally. The system updates trigger states from the scores and groups consecutive active observations into intervals.

\noindent\textbf{\textit{Multi-Condition Training with Shared Frames.}}
A training example consists of continuous frames and several monitoring conditions. Each condition has temporal labels indicating whether it holds at each step. A single causal forward pass predicts scores across time steps and conditions, and each score is compared with its label. The frames appear only once, allowing conditions to share visual representations and observation history. Losses are computed independently for each condition, so conditions with overlapping intervals can be trained together. Each training group contains at most eight conditions. Larger sets are split into groups.

We adapt Qwen3.5 with a frozen visual encoder and LoRA~\citep{hu2021lora} updates to the language model. The objective combines per-step judgments with interval supervision to improve both individual predictions and their agreement with annotated intervals.

Predictions near the start and end of a condition affect trigger timing, so we assign them higher training weights. Let $p_{t,k}$ be the score for condition $c_k$ at time $t$, $y_{t,k}\in\{0,1\}$ its label, and $m_{t,k}\in\{0,1\}$ a validity mask. We compute weighted binary cross-entropy over valid positions as
\begin{equation}
\mathcal{L}_{\mathrm{CE}} =
\frac{\sum_{t,k}m_{t,k}w_{t,k}\,
\operatorname{BCE}(p_{t,k},y_{t,k})}
{\sum_{t,k}m_{t,k}w_{t,k}}.
\label{eq:robobeats-ce}
\end{equation}
Here, $\operatorname{BCE}$ measures the difference between a score and its label, and $w_{t,k}$ controls the weight of each position. Positions near interval boundaries receive weight three, and other valid positions receive weight one. This weighting emphasizes changes at condition onsets and offsets. Invalid positions are excluded by $m_{t,k}$.

\begin{figure}[t]
\centering
\includegraphics[width=\textwidth]{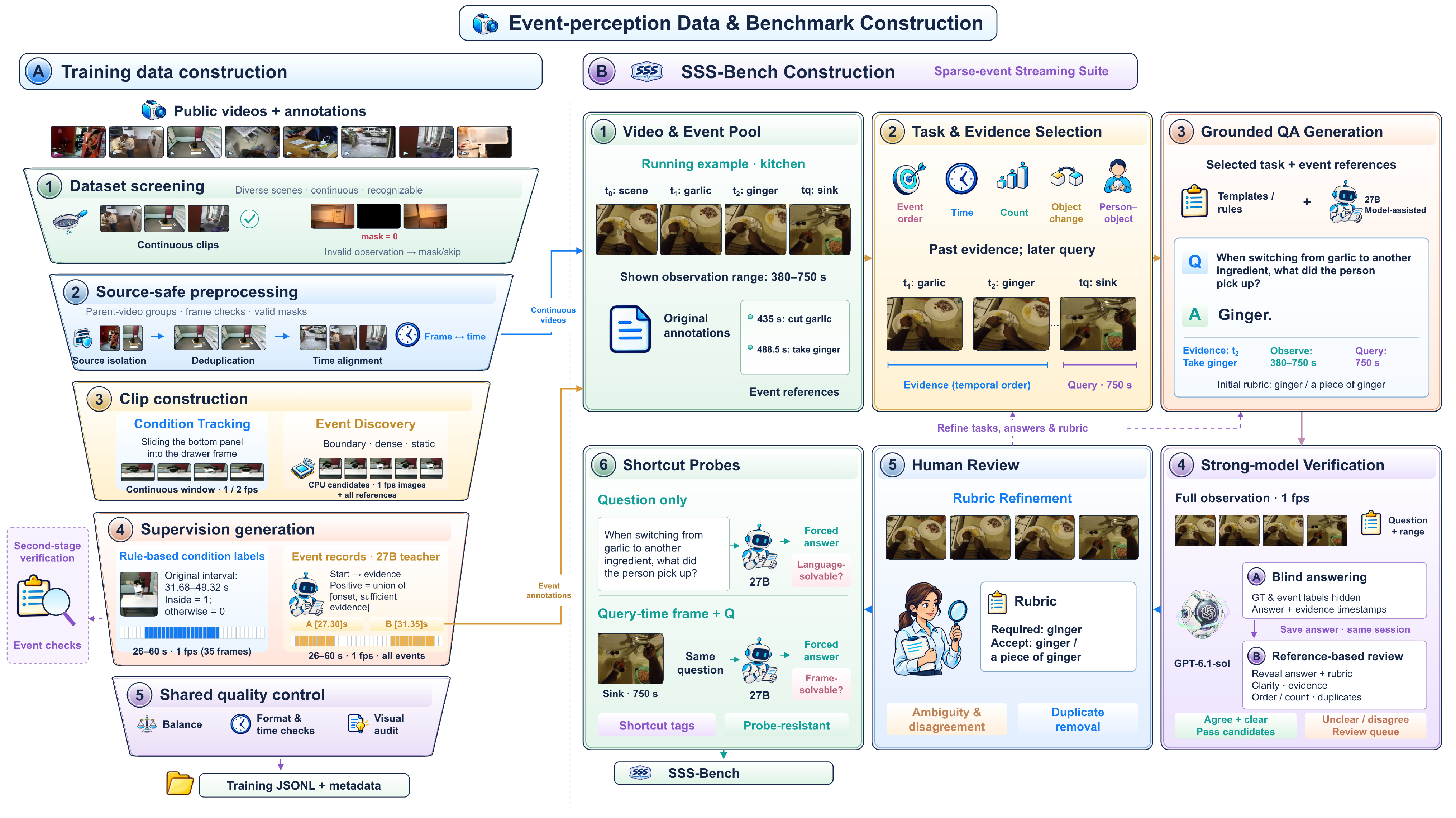}
\caption{\textbf{Construction of perception training data and SSS-Bench.} \textbf{(A) Perception training data.} Public videos and annotations are screened, grouped by source, deduplicated, and temporally aligned to form continuous clips. Condition-tracking labels follow native action intervals. For event discovery, a teacher identifies windows from event onset to sufficient visual evidence, which are merged into positive intervals. Both branches undergo data balancing, format and timestamp checks, and visual inspection. \textbf{(B) SSS-Bench.} Video evidence supports question construction, verification, and shortcut probing. The kitchen example relates historical evidence to a later question.}
\label{fig:robobeats-data}
\end{figure}

We also use a soft intersection-over-union loss to supervise positive intervals across the full prediction sequence. For the set $\mathcal{K}_{+}$ of conditions with valid positive labels, the loss is
\begin{equation}
\mathcal{L}_{\mathrm{IoU}} =
\frac{1}{|\mathcal{K}_{+}|}\sum_{k\in\mathcal{K}_{+}}
\left(1-
\frac{\sum_t m_{t,k}p_{t,k}y_{t,k}}
{\sum_t m_{t,k}(p_{t,k}+y_{t,k}-p_{t,k}y_{t,k})+\epsilon}
\right).
\label{eq:robobeats-iou}
\end{equation}
The constant $\epsilon$ provides numerical stability. This loss is zero when $\mathcal{K}_{+}$ is empty. The final objective is
\begin{equation}
\mathcal{L}=\mathcal{L}_{\mathrm{CE}}+
\lambda_{\mathrm{IoU}}\mathcal{L}_{\mathrm{IoU}},
\label{eq:robobeats-loss}
\end{equation}
where $\lambda_{\mathrm{IoU}}$ controls the weight of interval supervision. The combined objective trains per-step condition judgments and improves agreement between consecutive predictions and annotated intervals.

\noindent\textbf{\textit{Online Processing of Continuous Frames.}}

At the default 1\,Hz input rate, RoboBeats processes timestamped frames causally and reuses cached attention and recurrent states to avoid recomputing the full history. Overlapping chunks bound the context length while preserving recent observations across chunk transitions. Appendix~\ref{app:robobeats-online} gives the chunk sizes and cache-update procedure.

\noindent\textbf{\textit{Updating Trigger States.}}
Per-step scores can fluctuate as the scene changes. To reduce repeated switching, the system maintains an independent binary state for each condition and uses separate activation and deactivation thresholds. Their default values are 0.5 and 0.3. Let $p_{t,k}$ be the score for condition $c_k$ at time $t$ and $s_{t,k}$ its updated state. All states start at zero. The update rule is given in Appendix~\ref{app:trigger-state}.
A state of one denotes an active condition, and zero denotes an inactive condition. Between the two thresholds, the previous state is retained. A transition from zero to one records an onset, and a transition from one to zero records an offset. Consecutive active steps form an observation interval. Multiple conditions can be active at once. Downstream modules can use the onset to trigger a response or collect frames throughout the interval.

\begin{figure}[!t]
\centering
\includegraphics[width=\linewidth]{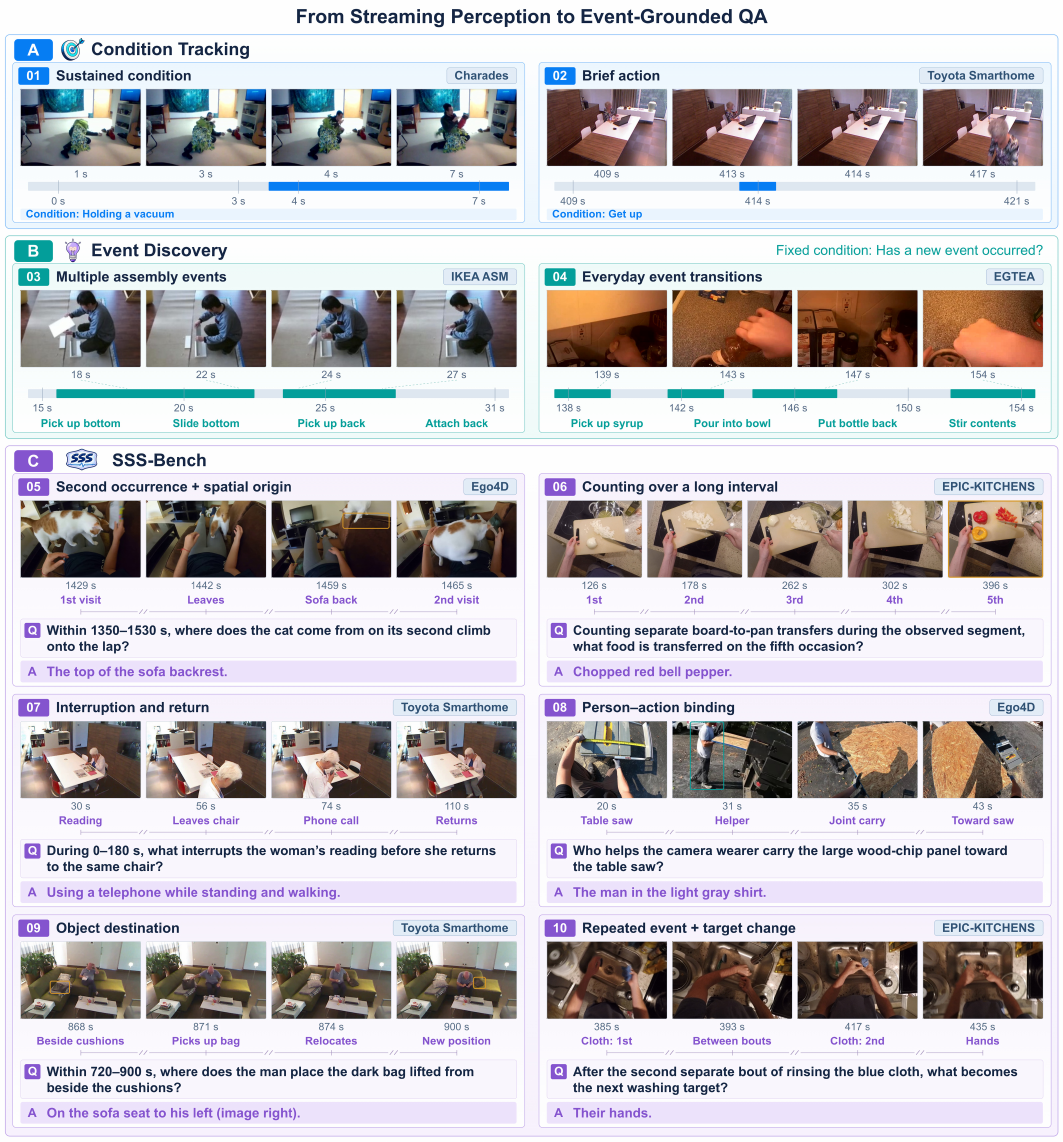}
\caption{\textbf{From streaming perception to event-grounded question answering.} (A) Condition tracking. (B) Event discovery. (C) SSS-Bench questions whose answers depend on earlier observations.}
\label{fig:data_overview}
\end{figure}

\subsection{Training Data Construction}
\label{sec:robobeats-data}

A robot observes the world as a continuous stream, tracking activities and changes as they unfold in its field of view. Edited videos may contain shot transitions that interrupt this continuity. To better match robot perception, we select temporally continuous recordings of everyday activities, including egocentric videos and videos in home environments. These observations provide the basis for condition tracking and event-discovery supervision.

\begin{figure}[t]
\centering
\includegraphics[width=\textwidth]{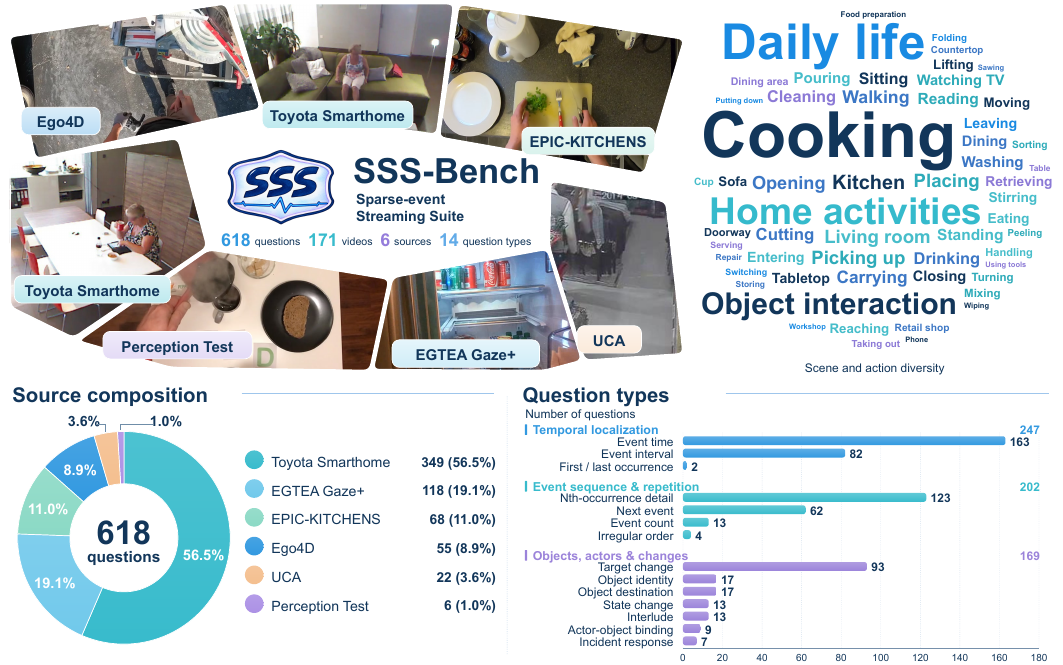}
\caption{\textbf{Overview of SSS-Bench.} The benchmark contains 618 questions over 171 parent videos from six sources. The panels show example frames, the scenes and actions covered, the source composition, and the number of questions in each of the three question families and 14 subtypes.}
\label{fig:sss_overview}
\end{figure}

We build training data from public videos and annotations to provide frame-level supervision for condition tracking and event discovery~\citep{sigurdsson2016charades,gao2017tall,dai2020tsu,benshabat2020ikea,grauman2021ego4d,damen2018epic,damen2020epic100,li2018egtea,patraucean2023perception,qasim2021edinburgh,yuan2023uca,epstein2019oops,kwolek2014urfall}. As shown in Figure~\ref{fig:robobeats-data}, construction comprises data screening, preprocessing, clip construction, label generation, and quality checks. We select temporally continuous clips with recognizable scenes and actions. Invalid frames are filtered or masked. We organize data splits by source video, remove duplicates, and align frames with timestamps.

For condition tracking, event descriptions and annotated intervals define continuous video windows containing the event and its surrounding frames. Frames within a condition's annotated interval are labeled one, and other valid frames are labeled zero. A clip can contain several monitoring conditions, each with independent frame-level labels. This matches the input format of multi-condition tracking.

For event discovery, we construct clips from continuous video and organize event records using native annotations and a 27B teacher model. A newly appearing activity or state change counts as a new event. If an activity is already underway at the beginning of a clip, the clip start is not treated as its onset. The end of an existing activity alone does not count as a new event. An activity resumed after an interruption can constitute a new event.

For each retained event $e$, let $a_e$ be its onset frame and $b_e$ the earliest frame with sufficient visual evidence. The interval from onset to sufficient evidence is positive. Taking the union across events gives the frame-level discovery labels
\begin{equation}
y^{\mathrm{disc}}_t = \mathbf{1}\!\left[t\in\bigcup_e[a_e,b_e]\right],
\label{eq:robobeats-discovery-label}
\end{equation}
where $t$ is the frame index and $\mathbf{1}[\cdot]$ is one when the enclosed condition holds and zero otherwise. The teacher uses the complete clip to generate supervision. The gate uses only the current and preceding frames during both training and online operation.

\begin{figure}[t]
    \centering
    \includegraphics[width=\linewidth]{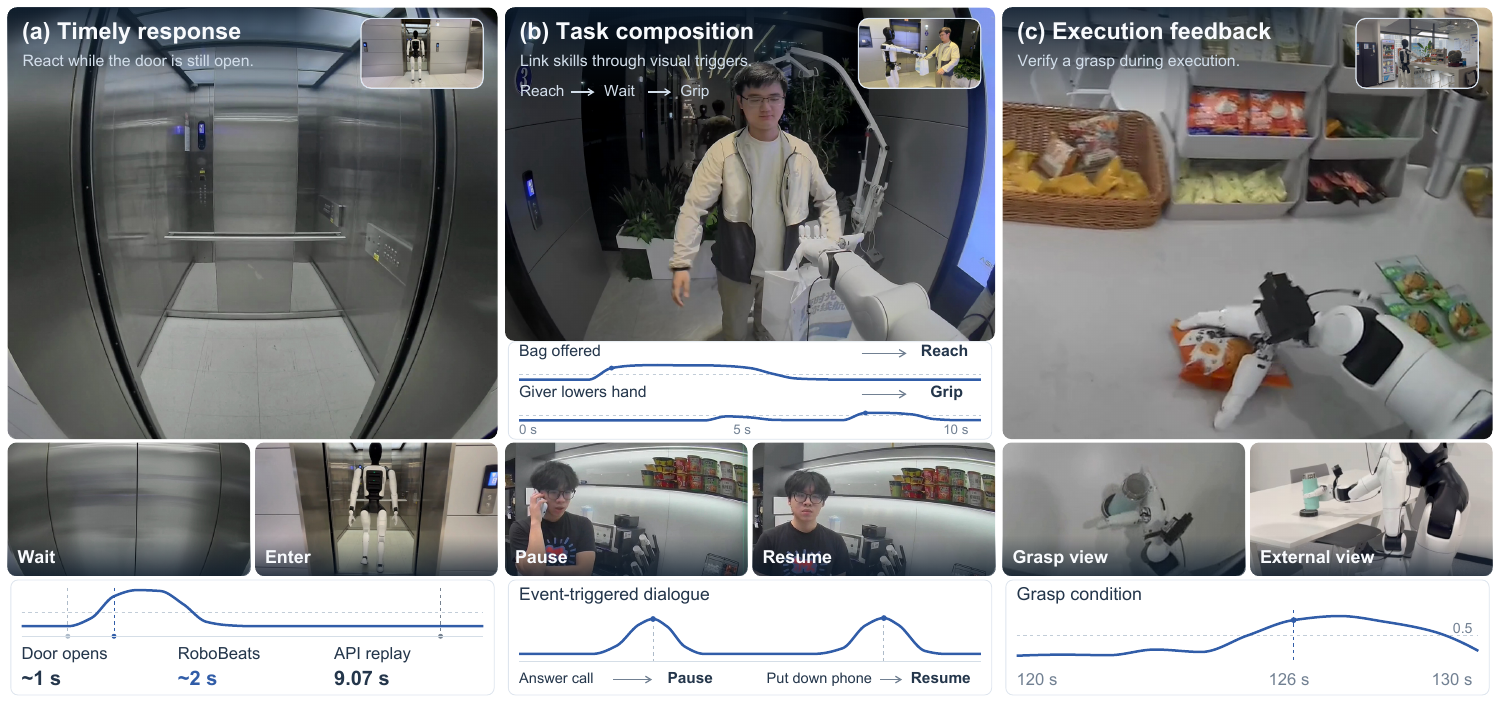}
    \caption{Three uses of condition tracking in robot interaction. (a) Timely response: the robot monitors an elevator door and enters when it opens; the API timing shown is from a replay of this example. (b) Task composition: visual conditions connect reach and grip during a bag handover, and trigger pauses and resumption during a phone call. (c) Execution feedback: the grasp condition is monitored while the robot grasps a snack.}
    \label{fig:robobeats-three-roles}
\end{figure}

We check data formats, temporal alignment, and valid observations, and inspect sampled frames to verify the supervision. The final data contain 31,068 multi-condition packs and 194,018 condition label sequences, together with 10,771 single-condition examples sampled from base conditions and 11,615 single-condition event-discovery examples.

\subsection{Benchmark Construction}
\label{sec:robobeats-benchmarks}

Existing online video benchmarks, such as OVO-Bench~\citep{li2025ovobench}, evaluate timely responses and questions about past events. Robots operating in everyday environments face an additional challenge: observations may continue for long periods, while activities and state changes are brief, sparse, and unevenly distributed. Processing every observation and retaining all details would increase computation and storage as experience accumulates. Long-term operation therefore calls for preserving potentially useful changes while filtering or compressing redundant observations.

We construct \textbf{SSS-Bench (Sparse-event Streaming Suite)} to study this setting using extended videos of everyday activities. The benchmark focuses on sparse events and their temporal, sequential, and object relations, while also including continuous activities and dense scenes for comparison. The system selects and organizes observations without knowing future questions, then receives a question at a specified time; all answer evidence precedes that time. Observation-selection strategies are compared under the same per-video writing-frame budget, so differences in question-answering accuracy reflect which observations each strategy retains. Figure~\ref{fig:data_overview} illustrates perception and memory question answering examples, while Figure~\ref{fig:sss_overview} summarizes the benchmark's sources and question types. SSS-Bench questions and reference answers are constructed from continuous videos and event cues, then checked through independent model verification and human review. Question-only and query-time-frame probes screen for potential shortcuts. Appendix~\ref{app:sss-construction-details} describes the construction and verification process.

The final benchmark contains \textbf{618 questions spanning 171 parent videos, six data sources, and 14 question subtypes}, with at most five questions per parent video. We check for overlap with the proactive gate's training set at four levels (videos, parent videos, person groups, and filenames) and find none. After independent verification, questions are selected according to predefined quotas for each question subtype. The selection is not adjusted based on the evaluation results of the methods being compared.

\subsection{From Perception to Responses and Memory}
\label{sec:robobeats-applications}

In LIFE, the agent or a predefined rule supplies conditions to RoboBeats, which tracks them as observations arrive. A change in a condition can prompt the system to respond or request further processing. The agent or task workflow determines the response from the gate output and the current context.

Figure~\ref{fig:robobeats-three-roles} illustrates three uses of condition tracking. At an elevator, the robot monitors the door and can enter while it is open. Visual conditions can also connect short actions: in the handover example, the robot reaches toward an offered bag and closes its hand when the giver lowers it. The same mechanism can pause speech when a person answers a phone call and resume it when the call ends.

Conditions can also be monitored while an action tool is running. In the snack-grasping example, RoboBeats tracks whether the grasp condition is met during execution. This observation provides feedback to the task workflow alongside the tool's own status, allowing the agent to assess the action in the context of the current scene. Similar monitoring could support the fetching stage of the assistance scenario in Figure~\ref{fig:life-experience-cases}.

Event discovery supports a different use of the same perception mechanism. It selects observations for a writer, which receives images and timestamps and describes visible events or states. The resulting records retain references to supporting frames and can be organized into Plastic Memory (Section~\ref{sec:life-memory}). Subsequent questions can retrieve the descriptions and return to the images when more detail is needed. Section~\ref{sec:real-robot-memory-qa} describes the offline construction of memory from a day of robot video.

\subsection{Experimental Evaluation}
\label{sec:robobeats-evaluation}

\subsubsection{Evaluation Tasks and Datasets}
We evaluate two capabilities of RoboBeats. With specified conditions, it must continuously judge whether they hold and trigger timely reminders. Without knowing future questions, it must discover events and select observations to retain as evidence for later question answering.

For condition tracking, we report condition judgment on held-out videos, with full results in Appendix~\ref{app:condition-heldout}, and evaluate on public benchmarks covering causal temporal localization (Charades-STA~\citep{gao2017tall}), first-trigger timing (StreamingBench proactive output~\citep{lin2024streamingbench}), action-boundary responses (SPOT-Bench~\citep{chatterjee2026spot}), and proactive responses with an answering model (ESTP-Bench~\citep{zhang2025estp}). Appendix~\ref{app:robobeats-eval-datasets} reports dataset sizes and the held-out set composition.

For event discovery, SSS-Bench tests whether observations selected before future questions are known support memory question answering. OVO-Bench Backward Tracing~\citep{li2025ovobench} provides a public comparison with the published OneStreamer results~\citep{zeng2026onestreamer}, covering episodic memory, action sequence identification, and hallucination detection.

\subsubsection{Experimental Setup and Baselines}
\label{sec:robobeats-evaluation-setup}
Unless otherwise specified, we evaluate RoboBeats adapted from Qwen3.5-2B. Input organization and online processing follow Sections~\ref{sec:robobeats-architecture} and~\ref{sec:robobeats-learning}.

Condition-tracking baselines include base models without gate adaptation, general VLMs~\citep{bai2025qwen25vl,bai2025qwen3vl,wang2025internvl35,yu2025minicpmv45,qwen2026qwen38}, and existing streaming models~\citep{wang2024mmduet,wang2025mmduet2,chang2025aha,zheng2026emgarde,yao2026joyai}. On the held-out set, external methods with multiple available readouts use the readout selected on the validation set. On Charades-STA, StreamingBench, and SPOT-Bench, general VLMs, the unadapted base models, and RoboBeats all use our condition-judgment protocol: the model judges at each step whether the condition holds, and a fixed readout converts these judgments into outputs. Inputs arrive at 1 fps, except on SPOT-Bench, where requests are rewritten as start and end conditions and judged at 2 fps, the input rate of the benchmark's streaming baselines. Published streaming methods are reported under their original protocols. For ESTP-Bench, a 27B model converts each question into monitoring conditions and generates an answer from recent frames after the gate triggers. Comparisons within this pipeline fix the answering model, prompts, and scoring procedure.

SSS-Bench observation selection comparisons fix the Qwen3.8-27B writer, reader, and scoring model, as well as the memory format and scoring procedure. In this simplified textual memory, selected frames are grouped into temporal packets; each packet is summarized independently, and the summaries are concatenated in chronological order. Under a per-video writing-frame budget, we compare RoboBeats, uniform sampling, MMDuet~\citep{wang2024mmduet} information-based triggering, and an event-discovery gate based on a Qwen3.8-27B model. We report two answering modes: text memory only, and text memory with one review window over the observed range that returns at most 10 frames. Figures~\ref{fig:sss-duration-results} and~\ref{fig:sss-type-results} use the latter. For OVO-Bench, RoboBeats-2B selects observations that a Qwen3.8-27B writer converts into textual memory. In module ablations, we compare this memory with memory constructed locally using OneStreamer PHCM, with Qwen3.8-27B and OneStreamer-4B as readers. For each reader, the corresponding answer-time inputs and scoring procedure are held fixed across memory sources.

\begingroup
\definecolor{blue}{HTML}{F1F6FC}\definecolor{blueink}{HTML}{173E5B}
\definecolor{mint}{HTML}{F1F8F4}\definecolor{mintink}{HTML}{234832}
\definecolor{slate}{HTML}{F3F5F8}\definecolor{slateink}{HTML}{354253}
\definecolor{cyan}{HTML}{EFF8FA}\definecolor{cyanink}{HTML}{1C4D59}
\definecolor{focus}{HTML}{D0EAF0}
\definecolor{roseink}{HTML}{2D7180}
\begin{table}[t]\centering
\caption{\textbf{Backward Tracing results on OVO-Bench (\%).} EPM: episodic memory; ASI: action sequence identification; HLD: hallucination detection. BT Avg. is the equally weighted mean of EPM, ASI, and HLD, except where noted; EPM+ASI Avg. averages EPM and ASI equally. Methods are grouped by how the systems operate. These groups are used for presentation and do not imply equal model sizes, training, or accessible evidence across methods.}\label{tab:ovo-backward-tracing}
\setlength{\tabcolsep}{3pt}\renewcommand{\arraystretch}{1.13}\small
\begin{tabularx}{\linewidth}{>{\raggedright\arraybackslash}X c *{4}{>{\centering\arraybackslash}m{12mm}} >{\centering\arraybackslash}m{27mm}}
\toprule
\textbf{Method} & \textbf{Size} & \textbf{EPM $\uparrow$} & \textbf{ASI $\uparrow$} & \textbf{HLD $\uparrow$} & \textbf{BT Avg. $\uparrow$} & \textbf{\shortstack{EPM+ASI\\Avg. $\uparrow$}} \\
\midrule
\rowcolor{blue}\multicolumn{7}{l}{\rule[-5.5pt]{0pt}{18pt}\textcolor{blueink}{\textbf{Proprietary MLLMs}}} \\
\rowcolor{blue} Gemini 1.5 Pro~\citep{google2024gemini15} & --- & 58.6 & 76.4 & 52.6 & 62.5 & 67.5 \\
\rowcolor{blue} GPT-4o~\citep{openai2024gpt4o} & --- & 57.9 & 75.7 & 48.7 & 60.8 & 66.8 \\
\addlinespace[3pt]
\rowcolor{mint}\multicolumn{7}{l}{\rule[-5.5pt]{0pt}{18pt}\textcolor{mintink}{\textbf{Open-Source Offline MLLMs}}} \\
\rowcolor{mint} Qwen2-VL-72B~\citep{wang2024qwen2vl} & 72B & 52.5 & 60.8 & 57.5 & 57.0 & 56.7 \\
\rowcolor{mint} LLaVA-Video-7B~\citep{zhang2024llavavideo} & 7B & 56.2 & 57.4 & 7.5 & 40.4 & 56.8 \\
\rowcolor{mint} Qwen3-VL-8B~\citep{bai2025qwen3vl} & 8B & 56.6 & 69.6 & 12.4 & 46.2 & 63.1 \\
\rowcolor{mint} Qwen3-VL-4B & 4B & 59.9 & 59.5 & 41.9 & 53.8 & 59.7 \\
\addlinespace[3pt]
\rowcolor{slate}\multicolumn{7}{l}{\rule[-5.5pt]{0pt}{18pt}\textcolor{slateink}{\textbf{Streaming Video MLLMs and Memory Methods}}} \\
\rowcolor{slate} MOSS-VL-Realtime~\citep{wang2026mossvl} & 11B & --- & --- & --- & 72.6 & --- \\
\rowcolor{slate} AURA~\citep{lu2026aura} & 8B & 54.9 & 67.6 & 58.6 & 60.4 & 61.3 \\
\rowcolor{slate} JoyAI-VL-Interaction~\citep{yao2026joyai} & 8B & 58.9 & 62.8 & 23.1 & 48.3 & 60.9 \\
\rowcolor{slate} Mage-VL~\citep{yang2026magevl} & 4B & 53.2 & 54.7 & 38.7 & 48.9 & 54.0 \\
\rowcolor{slate} VideoChat3~\citep{li2026videochat3} & 4B & 56.9 & 58.1 & 47.3 & 54.1 & 57.5 \\
\rowcolor{slate} Flash-VStream-7B~\citep{zhang2025flashvstream} & 7B & 39.1 & 37.2 & 5.9 & 27.4 & 38.1 \\
\rowcolor{slate} Dispider~\citep{qian2025dispider} & 7B + 1.5B & 48.5 & 55.4 & 4.3 & 36.1 & 51.9 \\
\rowcolor{slate} ReKV~\citep{di2025rekv} + STC\textsuperscript{a} & 7B & 50.5 & 55.4 & 30.1 & 45.3 & 53.0 \\
\rowcolor{slate} StreamForest-7B~\citep{zeng2025streamforest} & 7B & 58.9 & 64.9 & 32.3 & 52.0 & 61.9 \\
\rowcolor{slate} OneStreamer (reported)~\citep{zeng2026onestreamer} & 4B & 62.6 & 71.6 & 99.5 & 77.9 & 67.1 \\
\addlinespace[3pt]
\rowcolor{cyan}\multicolumn{7}{l}{\rule[-5.5pt]{0pt}{18pt}\textcolor{roseink}{\textbf{Video Agents}}} \\
\rowcolor{cyan} StreamAgent~\citep{yang2025streamagent} & 7B + 3B & 54.8 & 58.1 & 25.8 & 41.7\textsuperscript{b} & 56.5 \\
\rowcolor{cyan} EventMemAgent~\citep{wen2026eventmemagent} & 8B + 4B & 59.6 & 71.0 & 43.6 & 58.0 & 65.3 \\
\rowcolor{cyan} DeltaStreamer~\citep{sun2026deltastreamer} & 27B & 54.2 & 79.1 & 69.9 & 67.7 & 66.6 \\
\addlinespace[3pt]
\rowcolor{focus} RoboBeats (Qwen3.8-27B reader) & 2B & 62.3 & 70.9 & 70.4 & 67.9 & 66.6 \\
\rowcolor{focus} RoboBeats (OneStreamer-4B reader) & 2B & 64.0 & 71.6 & 99.5 & 78.4 & 67.8 \\
\bottomrule
\end{tabularx}
\par\vspace{3pt}{\footnotesize\raggedright
\noindent\textsuperscript{a} ReKV + STC uses both STC-Cacher and STC-Pruner; results are cited from the STC paper~\citep{wang2025stc}.
\par\noindent\textsuperscript{b} StreamAgent BT Avg. is reported as 41.7 in the original paper; the arithmetic mean of its three task scores is 46.2.
\par}
\end{table}
\endgroup

\subsubsection{Condition Tracking Evaluation}
\label{sec:robobeats-tracking-results}

\paragraph{Held-out condition judgment.}
On held-out videos, RoboBeats-2B achieves 0.915 AUROC and 73.6\% recall, compared with 0.612 and 14.1\% for its unadapted base. Full results are reported in Appendix~\ref{app:condition-heldout} (Table~\ref{tab:condition-heldout}).

\paragraph{Temporal localization and condition triggering.}
Charades-STA evaluates causal online interval localization using R@0.5; StreamingBench measures whether the first trigger falls within two seconds of the reference time. On Charades-STA, RoboBeats-2B reaches 57.3\%, compared with 7.4\% for its base. On StreamingBench, the corresponding scores are 56.0\% and 22.8\% (Table~\ref{tab:condition-public-main}; base-model results in Appendix~\ref{app:condition-full}).

\begin{table}[t]
\centering
\caption{\textbf{Results on three condition-tracking benchmarks (\%).} Metrics are causal online R@0.5 (Charades-STA), first-trigger accuracy within $\pm2$\,s (StreamingBench PO), and action-boundary T-F1@5 (SPOT-Bench). Published methods use their original protocols; general VLMs and RoboBeats use our condition-judgment protocol. RoboBeats training includes Charades-STA training data. Bold marks the best among our evaluated models with at most 8B parameters; dashes indicate unreported results. Appendix~\ref{app:condition-full} provides protocol details, full results, and untuned baselines.}

\label{tab:condition-public-main}
\small
\setlength{\tabcolsep}{5pt}
\renewcommand{\arraystretch}{1.12}
\definecolor{condGroup}{HTML}{F1F3F5}
\definecolor{condFocus}{HTML}{E3EDF8}
\begin{tabular}{lcccc}
\toprule
 \textbf{Method} & \textbf{Size} & \textbf{\shortstack{Charades-STA\\R@0.5}} & \textbf{\shortstack{StreamingBench\\PO $\pm2$s}} & \textbf{\shortstack{SPOT-Bench\\T-F1@5}} \\
\midrule
\rowcolor{condGroup}\multicolumn{5}{l}{\textbf{Streaming methods, published results, original protocols}} \\
StreamingVLM~\citep{chatterjee2026spot} & 7B & -- & -- & 3.9 \\
LiveStar~\citep{yang2026proagent} & 8B & -- & 13.6 & -- \\
Dispider~\citep{qian2025dispider} & 7B+1.5B & -- & 25.3 & -- \\
MMDuet2~\citep{wang2025mmduet2} & 3B & -- & -- & 14.2 \\
JoyAI-VL-Interaction & 8B+4B & -- & -- & 17.0 \\
Ctrl-SPOT~\citep{chatterjee2026spot} & -- & -- & -- & 22.2 \\
ProAgent (StreamingJev)~\citep{yang2026proagent} & 4B & -- & 57.2 & -- \\
\midrule
\rowcolor{condGroup}\multicolumn{5}{l}{\textbf{General VLMs, condition-judgment protocol}} \\
InternVL3.5 & 8B & 31.0 & 39.2 & 33.3 \\
MiniCPM-V-4.5 & 8B & 28.9 & 52.8 & 37.7 \\
Qwen2.5-VL & 7B & 26.6 & 45.6 & 42.3 \\
\midrule
\rowcolor{condFocus}RoboBeats-0.8B & 0.8B & 53.1 & 51.2 & 40.7 \\
\rowcolor{condFocus}RoboBeats-2B & 2B & \textbf{57.3} & 56.0 & 43.8 \\
\rowcolor{condFocus}RoboBeats-4B & 4B & 57.2 & \textbf{64.0} & \textbf{45.8} \\
\bottomrule
\end{tabular}
\end{table}

\paragraph{Timely responses at action boundaries.}
On SPOT-Bench, RoboBeats-2B achieves 43.8 T-F1@5, compared with 27.5 for its base under the same condition rewriting and output readout (Table~\ref{tab:condition-public-main}; base-model results in Appendix~\ref{app:condition-full}). Appendix~\ref{app:spot-readout} describes the output construction and its effect on comparisons with the official protocol. Figure~\ref{fig:robobeats-evaluation}(a--c) summarizes the three benchmarks.

\paragraph{Proactive responses with an answering model.}
On ESTP-Bench, the gate triggers a 27B answering model. Under the same answering pipeline, RoboBeats-2B achieves mean ESTP-F1 scores of 33.6 and 40.2 on explicit and implicit tasks, compared with 19.5 and 34.4 for uniform triggering (Table~\ref{tab:estp}; Figure~\ref{fig:robobeats-evaluation}(e)). Uniform triggering receives the same number of answering-model calls, 10.4 per question; RoboBeats raises 3.0 false alarms per question, compared with 2.7 for uniform triggering.
\definecolor{group0}{HTML}{F1F6FC}\definecolor{ink0}{HTML}{173E5B}
\definecolor{group1}{HTML}{F1F8F4}\definecolor{ink1}{HTML}{234832}
\definecolor{focus}{HTML}{D7ECDD}
\begin{table}[t]
\centering
\caption{\textbf{Condition gating and proactive response results on ESTP-Bench.} Bold marks the best result within the triggering-strategy comparison. For context question answering and overall results, $\dagger$ denotes values cited from the original papers.}\label{tab:estp}
\setlength{\tabcolsep}{1.4pt}
\renewcommand{\arraystretch}{1.15}
\footnotesize
\resizebox{\linewidth}{!}{%
\begin{tabular}{l*{18}{c}}
\toprule
\multirow{2}{*}{\textbf{Method}} & \multicolumn{9}{c}{\textbf{Explicit proactive tasks}} & \multicolumn{5}{c}{\textbf{Implicit proactive tasks}} & \multicolumn{3}{c}{\textbf{Context QA}} & \multirow{2}{*}{\textbf{Overall}} \\
\cmidrule(lr){2-10}\cmidrule(lr){11-15}\cmidrule(lr){16-18}
 & \textbf{OR} & \textbf{AP} & \textbf{TRU} & \textbf{OL} & \textbf{OSC} & \textbf{EOL} & \textbf{EOSC} & \textbf{AR} & \textbf{Avg.} & \textbf{OFR} & \textbf{IFR} & \textbf{NAR} & \textbf{TU} & \textbf{Avg.} & \textbf{ORC} & \textbf{TRC} & \textbf{Avg.} & \textbf{} \\
\midrule
\rowcolor{group0}\multicolumn{19}{l}{\textcolor{ink0}{\textbf{Existing Methods}}} \\
\rowcolor{group0} MMDuet & 6.7 & 9.6 & 17.1 & 9.8 & 3.8 & 5.6 & 8.6 & 7.5 & 8.6 & 9.5 & 7.3 & 48.0 & 68.5 & 33.3 & 17.4\textsuperscript{\textdagger} & 23.1\textsuperscript{\textdagger} & 20.3\textsuperscript{\textdagger} & 17.8\textsuperscript{\textdagger} \\
\rowcolor{group0} Qwen2-VL & 12.6 & 12.0 & 14.8 & 28.3 & 7.1 & 14.2 & 15.8 & 10.1 & 14.4 & 16.5 & 18.7 & 54.5 & 61.0 & 37.7 & 13.0\textsuperscript{\textdagger} & 7.7\textsuperscript{\textdagger} & 10.4\textsuperscript{\textdagger} & 21.3\textsuperscript{\textdagger} \\
\rowcolor{group0} MiniCPM-V & 13.2 & 15.5 & 16.2 & 24.9 & 6.5 & 11.7 & 11.6 & 12.3 & 14.0 & 14.3 & 20.0 & 44.2 & 60.9 & 34.8 & 24.6 & 27.8 & 26.2 & 22.9\textsuperscript{\textdagger} \\
\rowcolor{group0} VideoLLM-EyeWO & 19.4 & 24.4 & 18.7 & 23.1 & 16.1 & 20.4 & 20.9 & 19.9 & 20.4 & 23.6 & 26.3 & 52.4 & 34.8 & 34.3 & 39.4 & 54.9 & 47.1 & 28.2 \\
\addlinespace[3pt]
\rowcolor{group1}\multicolumn{19}{l}{\textcolor{ink1}{\textbf{Triggering Strategy Comparison}}} \\
\rowcolor{group1} Uniform triggering & 17.3 & 20.1 & 20.7 & 33.9 & 13.2 & 16.8 & 19.8 & 14.0 & 19.5 & 22.2 & 20.6 & 41.3 & 53.5 & 34.4 & --- & --- & --- & --- \\
\rowcolor{group1} Qwen3.5-2B & 23.8 & 23.3 & 26.1 & 23.4 & 16.3 & 17.2 & 18.8 & 15.2 & 20.5 & 20.7 & 12.9 & 40.9 & 50.7 & 31.3 & --- & --- & --- & --- \\
\rowcolor{focus} RoboBeats-2B & \textbf{24.1} & \textbf{34.7} & \textbf{43.0} & \textbf{39.3} & \textbf{32.1} & \textbf{28.2} & \textbf{41.0} & \textbf{26.0} & \textbf{33.6} & \textbf{30.6} & \textbf{21.5} & \textbf{49.0} & \textbf{59.7} & \textbf{40.2} & 47.3 & 30.8 & 39.1 & 36.2 \\
\bottomrule
\end{tabular}%
}
\end{table}

\subsubsection{Event Discovery Evaluation}
\label{sec:robobeats-discovery-results}

\begin{figure}[!t]
\centering
\includegraphics[width=0.90\linewidth]{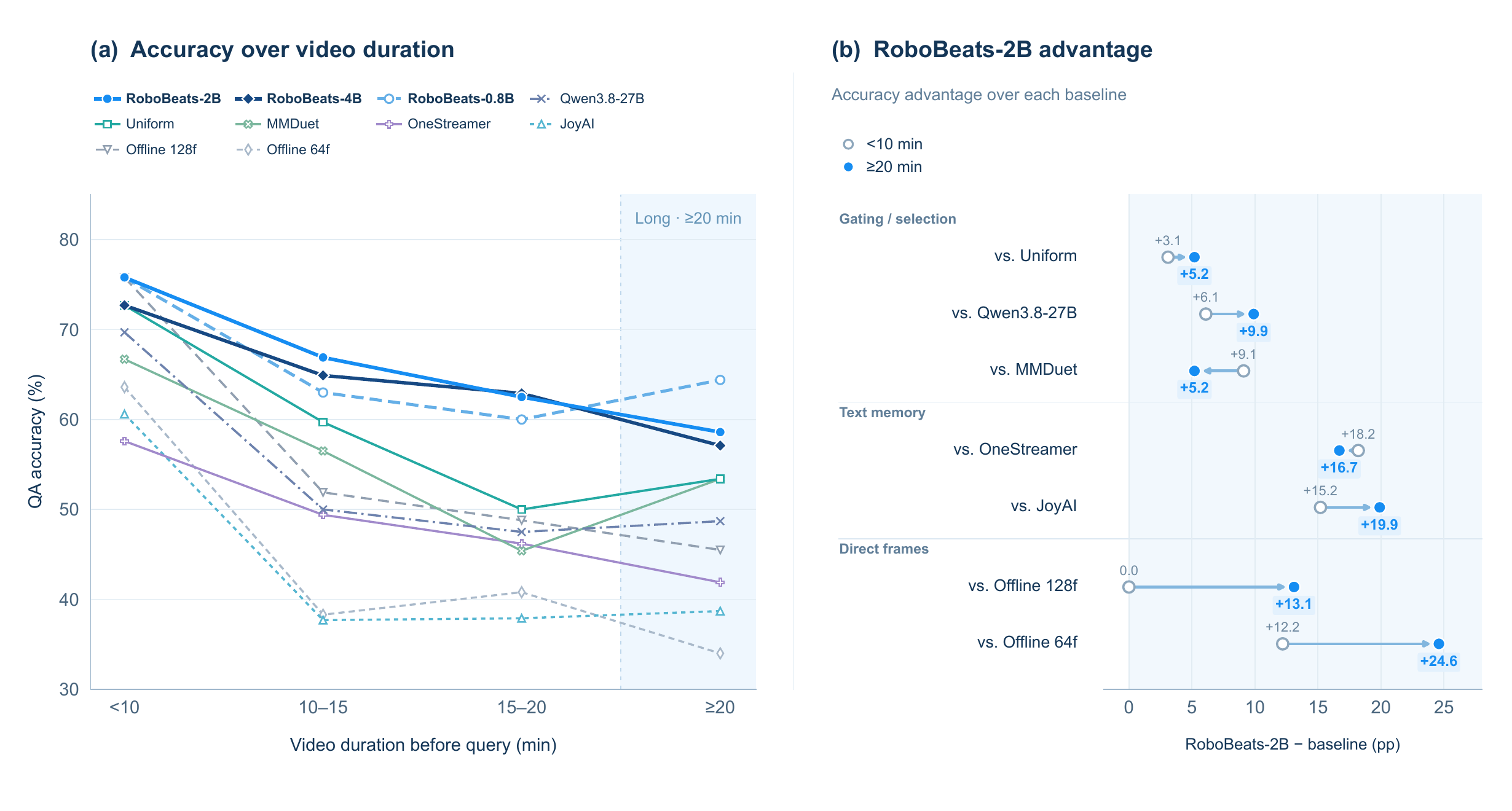}
\caption{\textbf{Memory question answering by observation duration on SSS-Bench.} (a) Accuracy across observation durations. (b) Accuracy differences between RoboBeats-2B and each baseline.}
\label{fig:sss-duration-results}
\end{figure}

\begin{figure}[t]
\centering
\includegraphics[width=0.64\linewidth,height=0.52\textheight,keepaspectratio]{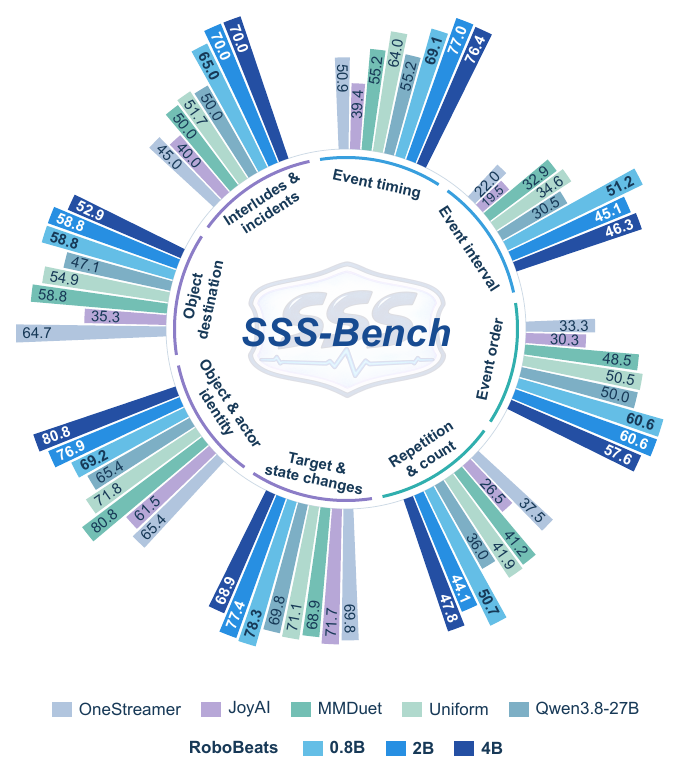}
\caption{\textbf{Memory question answering accuracy by question type on SSS-Bench.}}
\label{fig:sss-type-results}
\end{figure}

\paragraph{Effects of observation selection on memory question answering.}
We analyze SSS-Bench accuracy by observation duration and question type. Observation duration is the time from the start of observation to the question. Comparisons of frame selection methods follow Section~\ref{sec:robobeats-evaluation-setup}. As supplementary comparisons, we pass memories generated by OneStreamer~\citep{zeng2026onestreamer} and JoyAI-VL-Interaction to the same Qwen3.8-27B reader and report offline question answering from 128 frames. As shown in Figure~\ref{fig:sss-duration-results}, accuracy generally decreases as the observation period grows. RoboBeats continues to retain useful evidence over longer periods. For observations of at least 20 minutes, RoboBeats-2B exceeds uniform sampling and the Qwen3.8-27B baseline by 5.2 and 9.9 percentage points, respectively.

Figure~\ref{fig:sss-type-results} reports results by question type. Compared with uniform sampling, RoboBeats-2B improves accuracy by 13.0 and 10.5 percentage points on event timing and event interval questions, respectively. Gains also extend to questions about event order, target changes, and state changes. These results indicate that event-discovery gating helps retain temporal cues and information about activity changes from past observations, providing useful evidence for subsequent question answering. Overall results and comparisons across answering modes are reported in Table~\ref{tab:memory-qa-writer-comparison} (Appendix~\ref{app:sss-additional}) and summarized in Figure~\ref{fig:robobeats-evaluation}(f).

\paragraph{Memory question answering on a public benchmark.}
We further evaluate the retained information on the OVO-Bench~\citep{li2025ovobench} Backward Tracing tasks. RoboBeats-2B selects observations, and a Qwen3.8-27B writer converts them into textual memory. We evaluate this memory with OneStreamer-4B and Qwen3.8-27B readers. Table~\ref{tab:ovo-backward-tracing} reports task accuracies, their equally weighted mean (BT Avg.), and the equally weighted EPM--ASI mean (EPM+ASI Avg.). With the OneStreamer-4B reader, RoboBeats memory achieves a BT Avg. of 78.4\%, compared with the published OneStreamer result of 77.9\%~\citep{zeng2026onestreamer}, a difference of 0.5 percentage points. Figure~\ref{fig:robobeats-evaluation}(d) uses this published comparison.

HLD scores require interpretation alongside the answer distribution. All 186 HLD questions inspected locally have the reference answer ``Unable to answer.'' OneStreamer-4B achieves 99.5\% using recent frames without memory. The HLD score on this subset is therefore sensitive to the reader's abstention behavior. We retain the official BT Avg. and also report EPM+ASI Avg. to assess the use of historical information. The table distinguishes literature results and our evaluated module configurations.

\begin{table}[t]
\definecolor{focus}{HTML}{E3EDF8}
\centering
\caption{\textbf{Text-memory QA across readers (\%).} Readers use only the question and memory under a fixed answering setup. RoboBeats-gated pairs a RoboBeats-2B gate with a Qwen3.8-27B writer. EPM+ASI Avg. is the two-task mean; Gain is relative to OneStreamer PHCM.}\label{tab:text-memory-cross-reader}
\setlength{\tabcolsep}{2pt}\renewcommand{\arraystretch}{1.15}\fontsize{8}{9.5}\selectfont
\begin{tabularx}{0.8\linewidth}{l X *{4}{c}}
\toprule
\textbf{Reader} & \textbf{Memory source} & \textbf{EPM $\uparrow$} & \textbf{ASI $\uparrow$} & \textbf{\shortstack{EPM+ASI\\Avg. $\uparrow$}} & \textbf{Gain (pp)} \\
\midrule
\multirow{2}{*}{\textbf{Qwen3.8-27B}} & OneStreamer PHCM & 41.1 & 59.5 & 50.3 & --- \\
& \cellcolor{focus}RoboBeats-gated & \cellcolor{focus}\textbf{46.8} & \cellcolor{focus}\textbf{66.2} & \cellcolor{focus}\textbf{56.5} & \cellcolor{focus}+6.2 \\
\midrule
\multirow{2}{*}{\textbf{OneStreamer-4B}} & OneStreamer PHCM & 44.8 & 59.5 & 52.1 & --- \\
& \cellcolor{focus}RoboBeats-gated & \cellcolor{focus}\textbf{48.1} & \cellcolor{focus}\textbf{68.2} & \cellcolor{focus}\textbf{58.2} & \cellcolor{focus}+6.1 \\
\bottomrule
\end{tabularx}\end{table}
 We next fix the reader and answer-time inputs and compare textual memories generated by RoboBeats and OneStreamer PHCM. When answers use only textual memory, RoboBeats improves EPM+ASI Avg. by 6.2 percentage points with the Qwen3.8-27B reader and 6.1 points with the OneStreamer-4B reader (Table~\ref{tab:text-memory-cross-reader}). The selected observations thus provide useful historical evidence for both readers. Additional comparisons across memory sources and readers are reported in Table~\ref{tab:memory-source-grouped} (Appendix~\ref{app:ovo-additional-experiments}).

\paragraph{Writing volume and question answering performance.}
\begin{table}[t]
\definecolor{focus}{HTML}{E3EDF8}
\centering
\caption{\textbf{Writing cost and memory QA.} Readers use text memory and 16 recent frames; the recent-frames baseline uses no memory. Writers: 27B = Qwen3.8-27B; 4B = OneStreamer-4B. QA scores are percentages; EPM+ASI Avg. is the two-task mean. Bold compares memory-construction methods within each reader.}\label{tab:writing-volume-performance}
\setlength{\tabcolsep}{2pt}\renewcommand{\arraystretch}{1.15}\fontsize{8}{9.5}\selectfont
\begin{tabularx}{0.8\linewidth}{l X c c c c c}
\toprule
\textbf{Reader} & \textbf{Memory source} & \textbf{\shortstack{Writer\\size}} & \raisebox{-4pt}[0pt][0pt]{\fontsize{7}{7}\selectfont\bfseries\shortstack{Writer input\\[-1pt](M tokens/h) $\downarrow$}} & \textbf{EPM} & \textbf{ASI} & \textbf{\shortstack{EPM+ASI\\Avg. $\uparrow$}} \\
\midrule
\multirow{4}{*}{\textbf{Qwen3.8-27B}} & All frames & 27B & 0.338 & \textbf{67.7} & \textbf{72.3} & \textbf{70.0} \\
& Most recent 16 frames & --- & 0 & 65.0 & 63.5 & 64.2 \\
& OneStreamer PHCM & 4B & 23.8 & 65.3 & 68.2 & 66.8 \\
& \cellcolor{focus}RoboBeats-gated & \cellcolor{focus}27B & \cellcolor{focus}\textbf{0.186} & \cellcolor{focus}62.3 & \cellcolor{focus}70.9 & \cellcolor{focus}66.6 \\
\midrule
\multirow{4}{*}{\textbf{OneStreamer-4B}} & All frames & 27B & 0.338 & \textbf{65.3} & 66.9 & 66.1 \\
& Most recent 16 frames & --- & 0 & 62.3 & 64.2 & 63.2 \\
& OneStreamer PHCM & 4B & 23.8 & 61.3 & 69.6 & 65.4 \\
& \cellcolor{focus}RoboBeats-gated & \cellcolor{focus}27B & \cellcolor{focus}\textbf{0.186} & \cellcolor{focus}64.0 & \cellcolor{focus}\textbf{71.6} & \cellcolor{focus}\textbf{67.8} \\
\bottomrule
\end{tabularx}\end{table}
 We compare writer input volume and question answering performance across memory construction methods in Table~\ref{tab:writing-volume-performance}. Relative to writing the full frame sequence, RoboBeats gating reduces input tokens to the Qwen3.8-27B writer by approximately 45\%. Relative to OneStreamer PHCM, writer input tokens decrease by approximately 99.2\%. With the Qwen3.8-27B reader, EPM+ASI Avg. remains similar (66.6\% versus 66.8\%). With the OneStreamer-4B reader, RoboBeats achieves an EPM+ASI Avg. of 67.8\%, exceeding OneStreamer PHCM by 2.4 percentage points and all-frame writing by 1.7 points. These results show that event-discovery gating retains evidence for subsequent questions with fewer writer input tokens by selecting which observations to write. Appendix~\ref{app:ovo-additional-experiments} (Table~\ref{tab:deltastreamer-robobeats}) also reports the accuracy and token costs of integrating RoboBeats with DeltaStreamer.

Condition tracking supports timely responses to conditions of interest, while event discovery selects observations for further processing and recording. The next section describes how Plastic Memory organizes retained observations and makes them available for later use.

\section{Plastic Memory: Turning Experience into Reusable Evidence}
\label{sec:life-memory}

An experience retained today may only reveal its value tomorrow. A desktop seen during a patrol or a brief interaction between a person and an object may initially appear incidental, yet later become essential to answering a question or finding a target. Long-term memory must therefore support both efficient access to known clues and reinspection of past observations for information that was not extracted when they were recorded.

Existing robot memory systems use spatial maps and scene graphs~\citep{resight_vlmaps,resight_conceptgraphs,resight_hovsg}, visual observations for later retrieval and localization~\citep{resight_raven,resight_mom}, and structured memories that can be updated through experience~\citep{liu2026betamem}. LIFE uses Plastic Memory, a flat memory built on retained observations. Structured fields provide access paths for anticipated questions, while the images remain available for reinterpretation when a later question needs information that was not extracted at recording time.

Plastic Memory adopts the flat, observation-based memory design of our earlier work ReSight~\citep{che2026resight} and applies it to experience that accumulates through robot operation. Section~3 describes how proactive perception selects observations for further processing and recording. This section presents memory representation, access, and extension, followed by ReSight's base retrieval evaluation. Section~\ref{sec:life-edm} introduces ReSight's EDM mechanism for constructing and reusing tools, skills, and memory builders.

\begin{figure}[t]
    \centering
    \includegraphics[width=\textwidth]{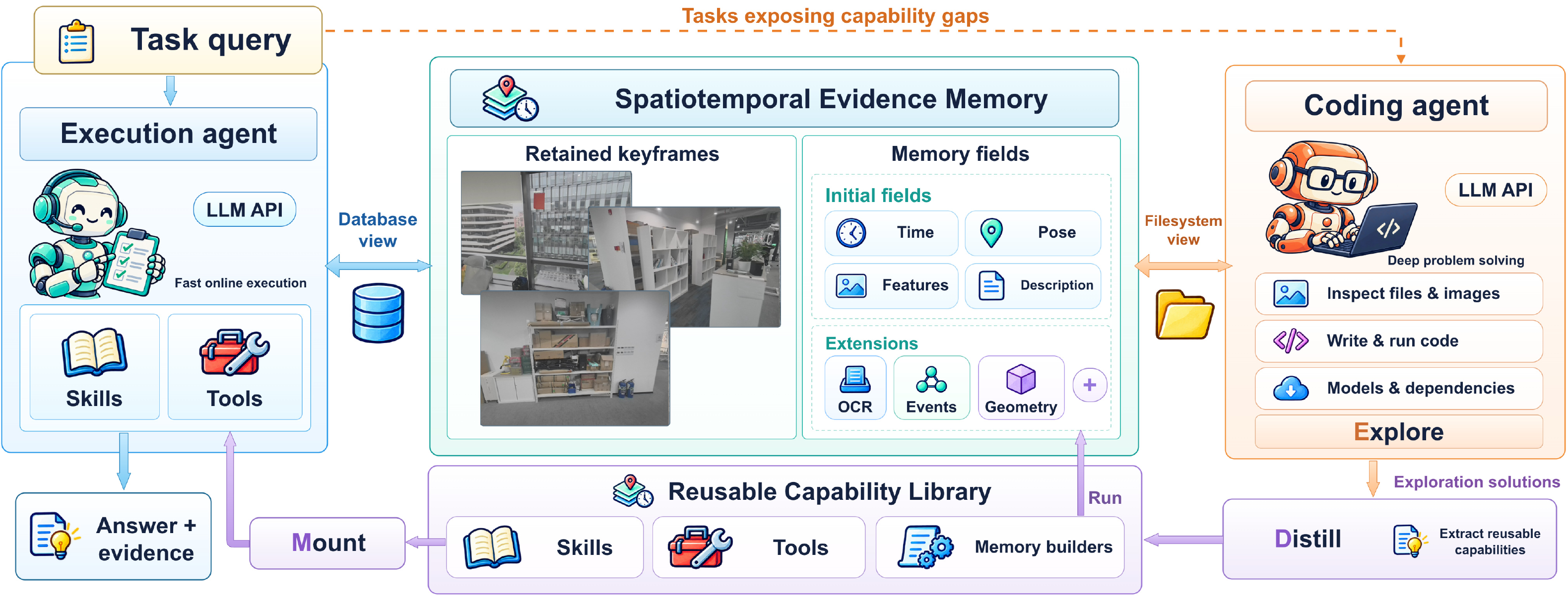}
    \caption{\textbf{Plastic Memory and capability expansion in LIFE.} Retained keyframes, initial fields, and extended fields form shared spatiotemporal evidence. The execution agent retrieves evidence through tools; the coding agent develops processing methods in a programmable workspace. EDM organizes exploration outputs into tools, skills, and memory builders for subsequent reuse. This diagram focuses on memory and capability expansion; the complete LIFE architecture is presented in Section~2.}
    \label{fig:life-memory-overview}
\end{figure}

As shown in Figure~\ref{fig:life-memory-overview}, an execution agent accesses shared evidence through tools, while a coding agent reprocesses observations through file and program interfaces. Both operate on the same retained evidence.

\subsection{A Reinterpretable and Programmable Memory Representation}
\label{sec:life-memory-evidence}

LIFE stores observations separately from their current interpretations. For each retained visual observation, the system preserves an image and its available context, including acquisition time, pose, and source. Semantic descriptions, people, objects, and events are organized as structured fields or associated records that provide access paths for later questions.

The memory is based on image files and associated structured records containing JSON fields; fields and derived indexes can be added or updated as needs change. A coding agent can inspect the images and write programs to read existing fields, extract new information, or construct additional indexes and tools. A field is a current interpretation of an observation, not all the information contained in it. For example, an image may initially be recorded with a location and object description. If a later task requires lookup by room number, retained images can be reprocessed to extract text and add a field or index linked to the source evidence.
\begin{equation}
\mathcal{M}^{(r)}=(\mathcal{O},\mathcal{F}^{(0)},\mathcal{X}^{(r)}),
\qquad o_i=(I_i,t_i,p_i,s_i).
\label{eq:life-memory}
\end{equation}
Here, $\mathcal{O}$ is the set of retained observations, and $I_i$, $t_i$, $p_i$, and $s_i$ denote the image, observation time, available pose, and source, respectively. Observations without poses remain accessible through visual and temporal retrieval. $\mathcal{F}^{(0)}$ contains the initial fields, and $\mathcal{X}^{(r)}$ contains derived fields and indexes after $r$ extensions. New fields change how historical experience can be accessed and interpreted, while the retained observations remain available as traceable evidence.

ReSight treats $\mathcal{O}$ as given. In LIFE, observations accumulate over time. Proactive perception can identify events or state changes that warrant further processing, after which the writing process selects and organizes relevant observations. We express the intended evidence-writing principle as
\begin{equation}
\mathcal{O}\leftarrow\mathcal{O}\cup\{o:\ g(o;\mathcal{M})>0\},
\label{eq:life-memory-write}
\end{equation}
where $g$ represents the potential information gain of an observation relative to existing memory. An observation may cover a previously unrecorded viewpoint or show new content at a covered viewpoint, such as a changed state or an event. In practice, source-specific selection and writing rules approximate $g$. Sections~3 and~\ref{sec:real-robot-memory-qa} describe proactive perception and the writing process in the 24-hour experiment.

Observations collected along a route and those selected by proactive perception can share a field schema while retaining their source information. Events can span multiple observations, so event descriptions retain references to supporting frames; other derived information likewise retains links to its evidence. Available fields depend on the source and writing process, and unavailable information can remain unset.

Raw evidence and derived fields are managed separately, with evidence references and retention policies supporting storage maintenance. Fields and indexes can thus be updated without losing traceability to their sources. Historical rescanning recovers details overlooked in retained images.

\subsection{Filtering and Linking Evidence for Question Answering and Action}
\label{sec:life-memory-access}

The same experience can be accessed from different directions. For ``Which cups are on the eighth floor?'', a floor field first narrows the candidates, followed by retrieval using object attributes. For ``Where did this person go afterwards?'', entity clues connect observations across time, and temporal ordering reconstructs the observed trajectory. Fields can both exclude irrelevant evidence and link otherwise scattered records. Let $\mathcal{R}$ denote memory records with available fields. We summarize these access patterns as
\begin{equation}
\begin{aligned}
\mathsf{Filter}_{P}(\mathcal{R})
  &=\{m_i\in\mathcal{R}:P(m_i)=1\},\\
\mathsf{Link}_{k}(S;\mathcal{R})
  &=\{m_j\in\mathcal{R}:\exists m_i\in S,\ R_k(m_i,m_j)=1\}.
\end{aligned}
\label{eq:life-filter-link}
\end{equation}
$P$ specifies a query constraint, and $R_k$ is a relation defined by a field or association rule, such as a shared entity identifier, a shared event, temporal proximity, or a common spatial region. Unknown attributes should remain distinguishable from known mismatches. Candidate identity and object associations also require inspection of the source evidence. Filtering and linking can be composed with similarity search, ranking, aggregation, and image inspection:
\begin{equation}
\mathcal{E}_q=(\mathsf{Op}_{L}\circ\cdots\circ\mathsf{Op}_{1})(\mathcal{M}),
\qquad y=\pi(q,\mathcal{E}_q).
\label{eq:life-evidence-query}
\end{equation}
$\mathcal{E}_q$ is the evidence selected for the current request, and $\pi$ uses it to produce an answer or inform an action. This is a common description of the access interface: individual operators may be fixed retrieval tools or programs executed on demand. Field-based association does not require every relation to be enumerated in advance. A query can combine conditions as needed, while recurring combinations can be packaged as tools.

For example, to ``find Peidong's cup'', the system may first retrieve identity or interaction records involving Peidong, inspect related sightings, and examine desktops or drinking events to connect person-related clues to visual evidence of the cup. A question can be answered with a candidate object and supporting evidence; a request to fetch it also takes the pose of the most recent reliable sighting, and the robot goes there as described in Section~\ref{sec:life-familiar}.

Historical evidence describes past states. When an object passes between people, scattered observations constrain its possible transfer history, but unobserved handovers remain inferences. Direct observations should remain distinguishable from conclusions inferred from clues. If the current holder or location is uncertain, asking a person or observing again can supply further evidence. Memory therefore supports questions that require no movement as well as tasks that require physical action.

\subsection{Linking Historical and Latest Observations by View}
\label{sec:life-memory-views}

Frames written under Equation~\eqref{eq:life-memory-write} enter one keyframe index, whether they were sampled along a route or saved by proactive perception when a state changed, an event occurred, a condition fired, or a person arrived. The source $s_i$ in Equation~\eqref{eq:life-memory} records how each observation was written; access and retention use the same fields for every source. Appendix~\ref{app:life-keyframes} details the fields and writing rules.

Spatial association organizes repeated observations of the same area. We define a \emph{view} as a local observation unit based on position and viewing direction: keyframes with nearby positions and directions, or nearby estimated gaze regions, can be grouped together. This geometric approximation supports retrieval of the history of a place across repeated sightings. For a set of existing views $\mathcal{V}$,
\begin{equation}
v(i)=\sigma(p_i;\mathcal{V}),
\qquad V_v=\{i:v(i)=v\}.
\label{eq:life-view-assignment}
\end{equation}
For a keyframe with an available pose, $\sigma$ selects the existing view whose anchor has the lowest matching cost among those satisfying the geometric criteria, or creates a new view if none matches. Views are maintained separately for each map. Frames without poses remain accessible through temporal and visual clues. View membership instantiates $R_k$ in Equation~\eqref{eq:life-filter-link}, linking observations of an area at different times. Appendix~\ref{app:life-view-geometry} gives the criteria and parameters.

Route-sampled and proactive-perception frames join views in the same way, so a view collects both what a place looks like and what happened there. Any frame that passes the quality check in Appendix~\ref{app:life-keyframes} can serve as a spatial frame of its view.

The most relevant image for a historical question may precede the latest observation. Spatial retrieval therefore scores keyframes and merges candidates by view, returning both the best query match and the latest observation for each view. When two frames meet the comparison conditions, image similarity provides an appearance-change cue:
\begin{equation}
\begin{aligned}
c(v)&=\arg\max_{i\in V_v}t_i,
& r_q(v)&=\arg\max_{i\in V_v}s(q,i),\\
\Delta_q(v)&=\mathbf{1}\!\left[\cos(\mathbf{z}_{r_q(v)},\mathbf{z}_{c(v)})<\tau\right].
\end{aligned}
\label{eq:life-view-retrieval}
\end{equation}
Here, $s(q,i)$ is the retrieval score, $c(v)$ is the latest observation, $r_q(v)$ is the best query match, $\mathbf{z}$ is an image embedding, and $\tau$ is a similarity threshold. A value of $\Delta_q(v)=1$ flags low image similarity and prompts inspection of the two frames; confirming the present state takes a new observation. Appendix~\ref{app:life-view-retrieval} specifies retrieval channels, embedding requirements, and the minimum comparison interval.

For example, ``Where was the cup before?'' may require an older matching frame, while a task that requires going there takes the pose of the latest observation as the revisit target (Section~\ref{sec:life-familiar}). Historical question answering and physical action thus share evidence while retaining different requirements for recency.

Each view keeps its earliest and latest observations together with representative frames chosen for visual coverage and temporal spread, and historical queries can retrieve a view's frames near requested times. Appendices~\ref{app:life-retention} and~\ref{app:life-view-history} give the retention policy and the matching rules.

\subsection{Rescanning History and Extending Fields on Demand}
\label{sec:life-memory-extension}

As experience accumulates, memory must accommodate both new observations and questions unforeseen at recording time. States and events are linked to observations through timestamps, frame references, and source information. These interpretations can be revised, while retained images provide evidence for reinspection. Updating memory therefore also includes extracting new information from old experiences.

For example, existing fields may retrieve images of doorways without supporting lookup by room number. The system can rescan retained images, extract text, and build an index so that the same historical evidence supports new queries. For a new field builder $b$,
\begin{equation}
\mathcal{X}_{b}=b(\mathcal{O}_{S}),\qquad
\mathcal{M}^{(r+1)}=(\mathcal{O},\mathcal{F}^{(0)},\mathcal{X}^{(r)}\cup\mathcal{X}_{b}),\qquad
\mathcal{O}_{S}\subseteq\mathcal{O}.
\label{eq:life-memory-backfill}
\end{equation}
$\mathcal{O}_{S}$ is the historical scope to process, such as a patrol, a floor, or candidate segments relevant to a target. Builder outputs keep their links to the source frames. Processing may run once or be repeated as relevant observations accumulate; each new question need not trigger a rescan of the entire history.

Field backfilling changes how historical evidence can be accessed. OCR fields make old images searchable by text, and a stable lookup interface makes these fields directly available to subsequent tasks. The fields LIFE adds on the robot fit the same equation. The people visible in a frame, registered object instances, events, and view membership are each produced by a builder over retained frames, and these builders form part of the capability set described in Section~\ref{sec:life-edm}. A builder written by hand and one produced by EDM play the same role: as in ReSight, initial and extended fields differ in when they are introduced, not in their source.

The next section introduces ReSight's EDM mechanism for constructing, validating, and retaining such methods in response to capability gaps exposed by tasks.

\subsection{Evaluation of Base Memory Retrieval}
\label{sec:life-memory-eval}

We report ReSight's base retrieval results on GOAT-core~\citep{resight_lagmemo,che2026resight}, covering 480 image, object, and language queries across four scenes. The reproduced configurations use a common localization backend and return at most five candidates. A query succeeds if at least one candidate lies within 1.5\,m of the target in the navigation plane. Table~\ref{tab:life-goat} reports SR@5.

\begin{table}[t]
\centering
\caption{\textbf{GOAT-core retrieval without EDM: SR@5 (\%).} Results reproduced from ReSight~\citep{che2026resight}. The literature rows were reported in MoM Table~I~\citep{resight_mom} and reproduced in ReSight; HOV-SG's overall score averages only its two supported query types. Qwen and Sol denote the memory-annotation models; Sol is GPT-5.6-sol. Retrieval-only configurations use no query-time VLM reasoning. Groups differ in provenance and configuration and are not all reruns under a single matched protocol.}
\label{tab:life-goat}
\small
\definecolor{ptsGroup}{HTML}{F1F3F5}
\definecolor{ptsFocus}{HTML}{E3EDF8}
\begin{tabular}{lrrrr}
\toprule
 \textbf{Method} & \textbf{Image $\uparrow$} & \textbf{Object $\uparrow$} & \textbf{Language $\uparrow$} & \textbf{Overall $\uparrow$} \\
\midrule
\rowcolor{ptsGroup}\multicolumn{5}{l}{\textbf{Literature results reported in ReSight}}\\
DynaMem~\citep{resight_dynamem} & 16.1 & 7.5 & 5.1 & 9.6 \\
VLMaps~\citep{resight_vlmaps} & 43.3 & 69.7 & 61.0 & 58.8 \\
LagMemo~\citep{resight_lagmemo} & 56.4 & 88.4 & 66.8 & 70.8 \\
OpenFunGraph~\citep{resight_openfungraph} & 53.4 & 67.1 & 69.0 & 63.1 \\
HOV-SG~\citep{resight_hovsg} & -- & 70.8 & 82.3 & 76.5 \\
MoM~\citep{resight_mom} & 80.1 & 79.5 & 82.3 & 80.6 \\
\midrule
\rowcolor{ptsGroup}\multicolumn{5}{l}{\textbf{Agent configuration}}\\
\rowcolor{ptsFocus}ReSight (agent) & 92.5 & 90.7 & 88.6 & 90.6 \\
\midrule
\rowcolor{ptsGroup}\multicolumn{5}{l}{\textbf{Retrieval only: no query-time VLM reasoning}}\\
MoM (visual-only) & 88.2 & 72.7 & 83.5 & 81.5 \\
\rowcolor{ptsFocus}ReSight (retrieval, Qwen) & 90.1 & 91.3 & 88.0 & 89.8 \\
\rowcolor{ptsFocus}ReSight (retrieval, Sol) & 90.1 & 90.1 & 92.4 & 90.8 \\
\bottomrule
\end{tabular}
\end{table}

The Sol-annotated retrieval-only configuration achieves 90.8\% overall SR@5 without query-time VLM reasoning. Adding an agent improves image and object queries but lowers language-query performance, yielding a similar overall score. Well-organized fields and retrieval interfaces thus already provide strong evidence recall, and agent reasoning can be added when a question needs it.

\section{EDM: From Solving Once to Reusing Later}
\label{sec:life-edm}

\begin{figure}[t]
    \centering
    \includegraphics[width=\textwidth]{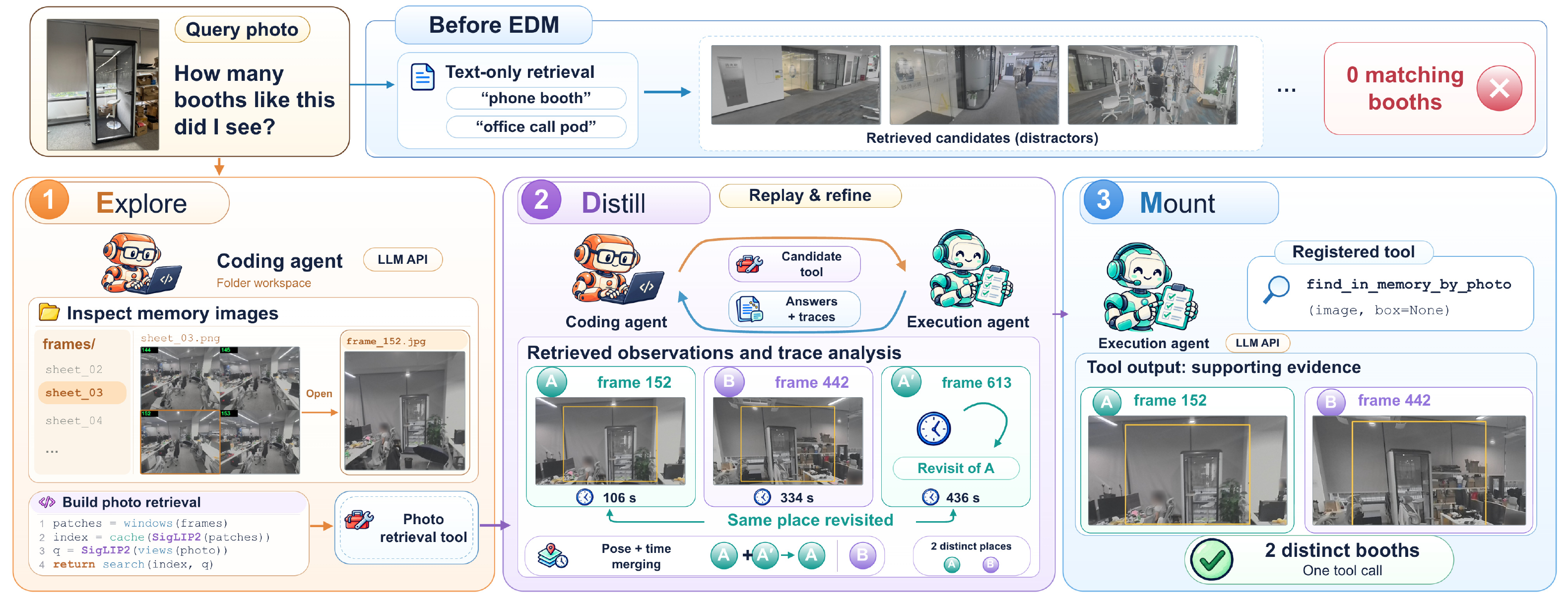}
    \caption{\textbf{Explore--Distill--Mount for a booth photo query in ReSight.} Explore inspects historical images and develops photo retrieval. Distill refines candidate tools using answers and traces from the execution agent, consolidating repeated sightings of the same place. Mount registers the tool and its interface. In the illustrated run, one call to the mounted tool returns two distinct booths and supporting frames. This call count is specific to the example; aggregate results appear in Table~\ref{tab:life-robot-memory}.}
    \label{fig:life-edm-case}
\end{figure}

The value of memory also depends on how effectively an agent can use it. Even when relevant evidence has been retained, the execution agent may lack an appropriate retrieval operation or require many interactions to align times, localize objects, or aggregate information across segments. A difficult task can therefore expose two different gaps: a missing fact, or a missing method for obtaining that kind of fact.

Prior work has explored reusable tools~\citep{resight_latm}, skill libraries for embodied agents~\citep{resight_voyager,resight_aspire}, and adaptation of memory systems~\citep{resight_memevolve}. Explore--Distill--Mount (EDM), introduced in our earlier work ReSight~\citep{che2026resight}, converts task-specific exploration into executable, verifiable, and reusable capabilities. This section describes EDM and reports its evaluation; all results here, including the robot-memory experiment, come from the original ReSight experiments rather than new LIFE evaluations. Section~6 shows an integrated LIFE system using memory and available tools to answer questions about a day of robot observations.

\subsection{Identifying Capability Gaps from Task Difficulty}

ReSight separates routine execution from capability construction. The execution agent answers queries using existing tools within a bounded online decision budget. The coding agent can directly inspect retained observations, write programs, invoke models, and construct intermediate artifacts to develop new solution paths for difficult tasks. These asymmetric interfaces allow complex analysis and development to take place separately while preserving a concise interface for routine execution.

For example, retrieval may identify images containing a target without returning the object's center in world coordinates. Segment descriptions may mention object pickups without reliably counting repeated interactions. Extending reasoning with the same tools does not necessarily supply these missing operations. EDM treats such requirements as opportunities to construct repeatable localization, counting, ordering, or retrieval programs.
\begin{equation}
\mathcal{C}=(\mathcal{T},\mathcal{S},\mathcal{B}),
\qquad (a,\tau)=\operatorname{Execute}(q,\mathcal{M};\mathcal{C}).
\label{eq:life-capabilities}
\end{equation}
$\mathcal{T}$ denotes callable tools, $\mathcal{S}$ denotes skills that guide their use, and $\mathcal{B}$ denotes builders that generate or update memory fields and indexes; $a$ and $\tau$ are the task output and execution trace. Adding a fact primarily changes memory, whereas retaining a method for obtaining facts changes the capability set. Self-evolution here operates at the level of programs, tool interfaces, and skills, without requiring base-model parameter updates after each task.

\subsection{Explore, Distill, and Mount}

During \textbf{Explore}, the coding agent inspects evidence for the current task, tests combinations of programs and models, and produces runnable solutions. It may extract text from relevant frames, estimate target locations using depth and pose, or organize events and revisit order. Exploration permits task-specific intermediate steps to identify missing operations and workable implementations.

During \textbf{Distill}, effective parts of the exploratory solution are converted into capabilities usable by the execution agent. This requires removing hard-coded scene paths and answers, specifying inputs, outputs, and applicability conditions, and refining tools and skill instructions using execution feedback. Distillation closes the gap between a coding agent solving a task once and a budget-limited execution agent invoking the same method reliably.

During \textbf{Mount}, validated tools, skills, and builders are integrated into the execution environment. Subsequent tasks access these capabilities through registered interfaces, with builders producing fields and indexes for the relevant memories when needed. For room-number lookup, constructing an OCR index processes evidence; registering a text-search tool and its usage instructions mounts a capability. Together, they make relevant evidence more directly accessible to future queries.
\begin{equation}
\begin{aligned}
Z&=\operatorname{Explore}(Q,\mathcal{M},\mathcal{C}),\quad \Delta\mathcal{C}=\operatorname{Distill}(Z;\operatorname{Execute}),\quad \mathcal{C}^{+}=\operatorname{Mount}(\mathcal{C},\Delta\mathcal{C}).
\end{aligned}
\label{eq:life-edm}
\end{equation}
$Q$ is the task set used for capability construction, and $Z$ contains exploration programs, evidence, and solution records. Equation~\eqref{eq:life-edm} summarizes the roles of the three stages; ReSight provides the complete algorithm, prompts, and implementation details~\citep{che2026resight}. We distinguish construction costs from subsequent use:
\begin{equation}
C_{\mathrm{total}}=C_{\mathrm{explore}}+C_{\mathrm{distill}}+C_{\mathrm{build}}
+\sum_{q\in Q_{\mathrm{use}}}C_{\mathrm{execute}}(q).
\label{eq:life-edm-cost}
\end{equation}
Construction is an upfront investment amortized over later tasks, so lower execution cost does not imply an equal reduction in total cost; the execution costs reported below cover only the last term.

Figure~\ref{fig:life-edm-case} follows EDM through the question ``How many booths like this reference photo have you seen?'' After text retrieval fails to find reliable matches, the coding agent inspects raw memory and builds photo retrieval. Replaying the execution agent then exposes duplicate counting caused by revisits, prompting consolidation of sightings using pose and time. Once the refined tool is registered, the execution agent can directly return two distinct booths with supporting evidence.

\subsection{Benchmark Evaluation of Capability Expansion}

We report two complementary ReSight evaluations. OC-NaVQA tests missing spatial and temporal operations; FindingDory tests event organization and queries across segments. Within each dataset, before/after EDM comparisons hold the execution model and online decision-round limit fixed to examine the effect of mounted capabilities.

\paragraph{OC-NaVQA.}
OC-NaVQA~\citep{resight_daaam} contains 210 queries over seven sequences. For position questions, the base system tends to return the observer's pose, whereas the task requires the target object's center. The EDM-derived operation combines target segmentation, depth backprojection, and multi-view aggregation while retaining source frames. Table~\ref{tab:life-ocnavqa} compares the same ReSight-Sol execution agent before and after mounting.

\begin{table}[t]
\centering
\caption{\textbf{OC-NaVQA results on 210 queries}, reproduced from ReSight~\citep{che2026resight}. Mini denotes GPT-5-mini, 2B/8B denote NVILA-Lite models, and Sol denotes GPT-5.6-sol. $\dagger$ marks configurations returning the observation-camera pose. The literature rows were reported by DAAAM~\citep{resight_daaam} and reproduced in ReSight. CodingAgent-Opus-5 uses a separate exploration budget and is shown as a reference, not a matched-budget online comparison. Bold marks the best result among the Sol execution configurations.}
\label{tab:life-ocnavqa}
\small
\setlength{\tabcolsep}{5pt}
\definecolor{ptsGroup}{HTML}{F1F3F5}
\definecolor{ptsFocus}{HTML}{E3EDF8}
\begin{tabular}{lrrrrr}
\toprule
 \textbf{Method} & \textbf{\shortstack{Binary (\%)\\$\uparrow$}} & \textbf{\shortstack{Text (\%)\\$\uparrow$}} & \textbf{\shortstack{Position error\\(m) $\downarrow$}} & \textbf{\shortstack{$\leq$10\,m (\%)\\$\uparrow$}} & \textbf{\shortstack{Time error\\(min) $\downarrow$}} \\
\midrule
\rowcolor{ptsGroup}\multicolumn{6}{l}{\textbf{Literature results reported in ReSight}}\\
ReMEmbR-2B~\citep{resight_remembr} & 43.2 & -- & 53.47 & -- & 2.287 \\
ReMEmbR-8B & 46.3 & -- & 55.89 & -- & 4.106 \\
ConceptGraphs~\citep{resight_conceptgraphs} & 29.9 & -- & 111.29 & -- & -- \\
DAAAM-Mini~\citep{resight_daaam} & 71.1 & -- & 41.75 & -- & 1.792 \\
\midrule
\rowcolor{ptsGroup}\multicolumn{6}{l}{\textbf{Sol execution configurations}}\\
DAAAM-Sol & 80.6 & 57.6 & 36.34 & \textbf{52.1} & 2.225 \\
RAVEN-Sol~\citep{resight_raven} $\dagger$ & 80.6 & 45.5 & 40.55 & 22.5 & 1.128 \\
ReSight-Sol (w/o EDM) $\dagger$ & \textbf{86.6} & \textbf{63.6} & 33.10 & 19.7 & 0.587 \\
\rowcolor{ptsFocus}ReSight-Sol (w/ EDM) & 82.1 & \textbf{63.6} & \textbf{24.72} & 36.6 & \textbf{0.475} \\
\midrule
\rowcolor{ptsGroup}\multicolumn{6}{l}{\textbf{Separate exploration-budget reference}}\\
CodingAgent-Opus-5 & 95.5 & 72.7 & 26.12 & 39.4 & 0.482 \\
\bottomrule
\end{tabular}
\end{table}

Position and time errors decrease after mounting, matching the added spatial and temporal operations, while text accuracy is unchanged and binary accuracy decreases: the gains concentrate on the operations that EDM constructed.

\paragraph{FindingDory.}
The main experiment evaluates FindingDory~\citep{yadav2025findingdory} on 596 queries from episodes 91--100, following RAVEN's long-horizon evaluation~\citep{resight_raven} and using official high-level success on spatial, temporal, and multi-goal questions. Both execution hosts use GPT-5.6-sol with a budget of 15 decision rounds per query. EDM exposes operations for interaction enumeration, revisit ordering, duration comparison, and temporal alignment. Table~\ref{tab:life-findingdory} reports the results.

\begin{table}[t]
\centering
\caption{\textbf{FindingDory capability construction, frozen reuse, and ablation}, combining two original ReSight tables~\citep{che2026resight}. Success is official high-level success (\%); Tokens/q is mean input-plus-output tokens per query in thousands. Gemini denotes Gemini-2.5-Flash and Sol denotes GPT-5.6-sol. The two groups use different episodes and are scored separately. EDM w/o Explore distills only from execution traces, excluding exploration programs, notes, and answers. Online tokens exclude exploration, distillation, and field-construction costs. Bold marks the best result among Sol execution configurations on the capability-construction episodes and among configurations on the held-out episodes.}
\label{tab:life-findingdory}
\small
\setlength{\tabcolsep}{4pt}
\definecolor{ptsGroup}{HTML}{F1F3F5}
\definecolor{ptsFocus}{HTML}{E3EDF8}
\begin{tabular}{lrrrrr}
\toprule
 \textbf{Method} & \textbf{Spatial $\uparrow$} & \textbf{Temporal $\uparrow$} & \textbf{Multi-goal $\uparrow$} & \textbf{Overall $\uparrow$} & \textbf{\shortstack{Tokens/q\\(k) $\downarrow$}} \\
\midrule
\rowcolor{ptsGroup}\multicolumn{6}{l}{\textbf{Capability-construction episodes: 91--100, 596 queries}}\\
\rowcolor{ptsGroup}\multicolumn{6}{l}{\textbf{Literature results from RAVEN, as reported in ReSight}}\\
Random & 13.4 & 6.9 & 0.5 & 9.2 & -- \\
Embedder Only & 31.0 & 8.1 & 3.4 & 18.8 & -- \\
VLM Only-Gemini & 36.0 & 28.1 & 2.2 & 28.2 & -- \\
ReMEmbR-Gemini~\citep{resight_remembr} & 36.7 & 17.1 & 4.5 & 25.0 & -- \\
RAVEN-Gemini~\citep{resight_raven} & 45.1 & 24.8 & 6.7 & 32.2 & -- \\
\addlinespace
\rowcolor{ptsGroup}\multicolumn{6}{l}{\textbf{Sol execution configurations}}\\
RAVEN-Sol & 51.9 & 34.8 & 5.6 & 38.9 & 378 \\
RAVEN-Sol (w/ EDM) & \textbf{55.9} & \textbf{89.5} & 48.3 & \textbf{66.6} & 150 \\
ReSight-Sol (w/o EDM) & 45.5 & 41.0 & 7.9 & 38.3 & 130 \\
\rowcolor{ptsFocus}ReSight-Sol (w/ EDM) & 54.2 & 82.9 & \textbf{50.6} & 63.8 & \textbf{43} \\
\addlinespace
\rowcolor{ptsGroup}\multicolumn{6}{l}{\textbf{Separate exploration-budget reference}}\\
CodingAgent-Opus-5 & 58.2 & 76.2 & 43.8 & 62.4 & -- \\
\midrule
\rowcolor{ptsGroup}\multicolumn{6}{l}{\textbf{Held-out episodes: 81--90, 597 queries; frozen capabilities}}\\
ReSight-Sol (w/o EDM) & 48.3 & 36.2 & 4.5 & 37.5 & 128.8 \\
ReSight-Sol (EDM w/o Explore) & 51.0 & 48.6 & 25.8 & 46.4 & \textbf{26.9} \\
\rowcolor{ptsFocus}ReSight-Sol (w/ EDM) & \textbf{57.4} & \textbf{83.3} & \textbf{52.8} & \textbf{65.8} & 39.4 \\
\bottomrule
\end{tabular}
\end{table}

EDM improves both hosts, with larger gains on temporal and multi-goal questions. This pattern is consistent with organizing scattered evidence into executable queries and shows applicability beyond one execution interface. Episodes 91--100 are used for exploration and distillation; the held-out episodes below test reuse.

On held-out episodes 81--90, which were excluded from exploration and distillation, we evaluate reuse of the frozen capabilities. Capabilities are not updated during evaluation. Mounting EDM increases overall high-level success from 37.5\% to 65.8\% and reduces tokens per query from 128.8k to 39.4k; the configuration without Explore uses fewer online tokens but reaches only 46.4\%, indicating that exploration outputs contribute to the reusable capabilities.

\subsection{Reusing Frozen Capabilities on Robot Memories}

We further evaluate whether constructed capabilities can support subsequent queries over memories recorded during office patrols. Capability construction uses only three tasks on eighth-floor memory. Tools and builders are then frozen and evaluated on six tasks, with five independent runs per task. Subsequent tasks include a new query about Room~805 and queries for patrol length and photo localization over first-floor memory collected on another day. Existing builders may run on the new memory, but capabilities are neither synthesized nor modified. Both configurations use the same execution model with at most ten tool-interaction rounds per run.

\begin{table}[t]
\centering
\caption{\textbf{Robot-memory query results}, reproduced from ReSight~\citep{che2026resight}. Each task is run independently five times. Tool calls and input-plus-output tokens are per-run means; the last row reports total successes and mean cost across 30 runs. Grouping distinguishes capability-construction tasks from subsequent tasks.}
\label{tab:life-robot-memory}
\small
\setlength{\tabcolsep}{5pt}
\definecolor{ptsGroup}{HTML}{F1F3F5}
\definecolor{ptsFocus}{HTML}{E3EDF8}
\begin{tabular}{lrrrrrr}
\toprule
& \multicolumn{2}{c}{\textbf{Success $\uparrow$}} & \multicolumn{2}{c}{\textbf{Tool calls $\downarrow$}} & \multicolumn{2}{c}{\textbf{Tokens/run (k) $\downarrow$}} \\
\cmidrule(lr){2-3}\cmidrule(lr){4-5}\cmidrule(l){6-7}
 \textbf{Task} & \textbf{w/o EDM} & \textbf{w/ EDM} & \textbf{w/o EDM} & \textbf{w/ EDM} & \textbf{w/o EDM} & \textbf{w/ EDM} \\
\midrule
\rowcolor{ptsGroup}\multicolumn{7}{l}{\textbf{A. Capability-construction tasks}}\\
Warmth Corner (8F) & 2/5 & \textbf{5/5} & 8.4 & \textbf{1.0} & 60.0 & \textbf{10.4} \\
Farthest point (8F) & 1/5 & \textbf{5/5} & 7.2 & \textbf{1.0} & 38.8 & \textbf{10.0} \\
Booth count (8F) & 0/5 & \textbf{5/5} & 7.6 & \textbf{1.0} & 50.2 & \textbf{12.9} \\
\midrule
\rowcolor{ptsGroup}\multicolumn{7}{l}{\textbf{B. Subsequent tasks: a new query (805) and new memory (1F)}}\\
Room 805 (8F) & \textbf{5/5} & \textbf{5/5} & 2.0 & \textbf{1.0} & 12.5 & \textbf{11.2} \\
Patrol length (1F) & 0/5 & \textbf{5/5} & 3.4 & \textbf{1.8} & \textbf{14.5} & 15.1 \\
Photo localization (1F) & 4/5 & \textbf{5/5} & 2.0 & \textbf{1.4} & \textbf{11.7} & 14.3 \\
\midrule
Total / mean & 12/30 & \textbf{30/30} & 5.10 & \textbf{1.20} & 31.29 & \textbf{12.31} \\
\bottomrule
\end{tabular}
\end{table}

With frozen capabilities, ReSight answers all 15 runs on subsequent tasks, compared with 9 of 15 before EDM (Table~\ref{tab:life-robot-memory}). Across the six tasks, successes increase from 12 of 30 to 30 of 30, while mean tool calls decrease from 5.10 to 1.20.

Building on these memory and tool designs, Section~\ref{sec:real-robot-memory-qa} presents an integrated robot system that uses a day's observations to answer questions about people, objects, and activities.

\section{A Day of Experience: 24-Hour Memory Question Answering on a Real Robot}
\label{sec:real-robot-memory-qa}

A robot that remains present in a shared environment can accumulate experiences before it knows what its users will ask. We bring together proactive perception, Plastic Memory, and the agent's tools to study how these experiences support later questions on an \textbf{A3 Ultra} robot operating in an office building. Over the day, its front-facing camera recorded work areas, a meeting room, and a pantry, as well as other floors of the building and outdoor areas, capturing arrivals, meetings, meals, and object handovers. We use footage from 21:00 on October 21 to 21:00 on October 22, 2026, to construct a memory of the day and ask the agent questions about the recorded experience.

\begin{figure}[p]
\centering
\includegraphics[width=\textwidth]{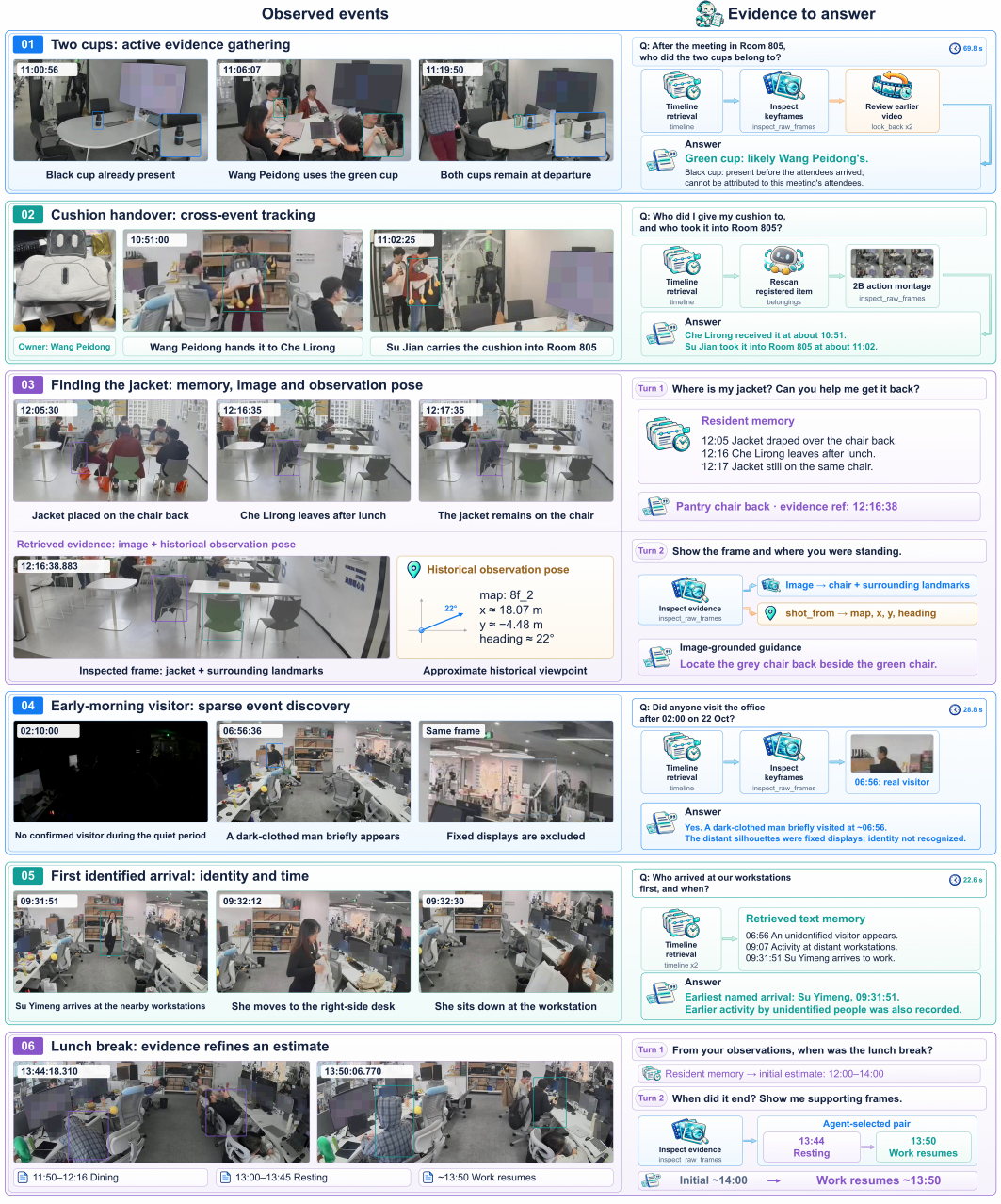}
\caption{\textbf{One day of observations and memory-grounded answers on A3 Ultra.} The six cases show cup attribution using additional visual evidence, a cushion handover linked across events, a jacket located using memory and a historical observation pose, an early-morning visitor, the first identified arrival at the workstations, and an estimate of the end of the lunch break refined with historical frames. The agent uses timestamped records and, where needed, revisits visual evidence to answer questions. The jacket question is asked as of 12:30, before its owner came back for it.}
\label{fig:robobeats-realrobot-cases}
\end{figure}

\subsection{From a Day of Observations to Memory}

We reconstruct memory offline from the recorded video, combining the proactive perception of Section~\ref{sec:active-perception} with Plastic Memory and its evidence-access design of Section~\ref{sec:life-memory}. The RoboBeats event-discovery gate scores incoming observations. Nearby triggers are combined into windows and passed through a background-difference prefilter before a local Qwen3.8-27B writer describes visible events and states. These descriptions retain timestamps and references to supporting frames, allowing a later question to lead back to the original observation.

A subsequent Qwen3.8-27B consolidation process organizes the writer's output in half-hour blocks, linking records of people, objects, places, and activities. Following the memory design in Section~\ref{sec:life-memory}, these descriptions and indexes provide access paths to timestamped visual evidence. Alongside the textual records, object detection and tracking provide additional observations of people and objects. For a registered object, condition-based rescanning can locate actions involving it across the recorded day. The agent uses tools for timelines, objects, and historical images to access these records. This follows the separation in Section~\ref{sec:life-edm}: evidence accumulates in memory, while tools and field builders make that evidence usable for different tasks.

Over the whole day, the gate and prefilter select 256 windows for the writer, and the resulting memory retains 1,296 keyframes; Appendix~\ref{app:real-robot-details} reports the full construction cost. This cost is paid once before any question is asked, and the same memory then serves every question without another pass over the video.

\subsection{Answering Questions from Experience}

The evaluation contains six questions about people, objects, times, and locations, one for each case in Figure~\ref{fig:robobeats-realrobot-cases}. Two of them, the jacket and the lunch break, are two-turn exchanges in which the user follows up on the first answer. LIFE's execution agent answers each question three times using the reconstructed memory and its tools, and the figure shows examples from these runs. The examples follow the path from selected observations to organized memory and tool-assisted answers: the agent can use a stored description, connect observations across events, and revisit a frame to examine evidence.

\paragraph{Tracing two cups through a meeting.}
A user asked who owned two cups left on a meeting-room table. The agent first queried the timeline and examined images of the cups after the meeting. It then used historical video access to review observations from earlier in the meeting and before attendees arrived. The green cup was associated with an attendee who brought and used it, while the black cup was already on the table before the attendees arrived. The agent therefore offered an attribution for the green cup and distinguished it from the black cup, whose owner was not established. This answer depended on the order in which the objects and people appeared, rather than on their proximity in the final scene.

\paragraph{Tracking a cushion across events.}
A user asked whom he had given his registered cushion to and who subsequently carried it into Room 805. The agent used the registered object and its memory records to identify the relevant cushion. It then used condition-based historical action evidence to connect an observed handover at approximately 10:51 with a later carrying event at approximately 11:02. It answered that Che Lirong received the cushion and Su Jian carried it into the room. The object record and historical action observations link the two questions to different people and times.

\paragraph{Finding a jacket.}
At 12:30, a user asked where the robot had last seen his jacket, then asked to see that observation and where the robot had been. The agent identified a chair back in the pantry, where the jacket was last recorded at about 12:18, showed the historical frame, and returned the map pose from which it had been taken. The memory record supplies the location, and the historical-image tool brings back the visual evidence and observation pose, giving the user a view of the place to search.

\paragraph{Inferring the end of a lunch break.}
Asked to estimate the lunch break from its observations, the agent combined records of dining, people resting, and subsequent activity at workstations, and answered approximately 12:00--14:00. When the user asked for a more precise end time and supporting images, it selected frames from approximately 13:41--13:50 and refined its answer: people began to resume work around 13:50. The exchange shows how a summary over several periods can guide inspection of images near a transition.

\paragraph{Other questions about the day.}
To answer whether anyone visited the office after 02:00, the agent found a brief visitor at approximately 06:56 and distinguished the person from fixed displays in the background. It also combined identity and time records to identify the first named arrival at the nearby workstations. These questions use the same day's experience as the object and activity examples, but access different fields and observations according to the information requested.

\subsection{Question-Answering Performance and Cost}

We compare LIFE's execution agent with two coding-agent configurations using the same GPT-6.1-sol answering model (Table~\ref{tab:real-robot-comparison}). The comparison uses the six questions above, each answered three times by every configuration. Codex A receives the recorded video, person and object registration materials, and poses. Codex B additionally receives the constructed memory. The LIFE configuration uses the robot's memory and tools. Each answer is manually scored against a rubric, with partial credit for some answers; Appendix~\ref{app:real-robot-details} gives the scoring and input details.

\begin{table}[t]
\centering
\caption{\textbf{Question answering on six real-robot cases.}}
\label{tab:real-robot-comparison}
\begingroup
\small
\definecolor{r17group}{HTML}{F1F3F5}
\definecolor{r17ink}{HTML}{3C4650}
\definecolor{r17focus}{HTML}{E3EDF8}
\newcommand{\rseventeenhead}[1]{#1}
\renewcommand{\tabularxcolumn}[1]{m{#1}}
\setlength{\tabcolsep}{5pt}
\renewcommand{\arraystretch}{1.18}
\begin{tabular}{lccc}
\toprule
\rseventeenhead{\textbf{Question / metric}} & \rseventeenhead{\textbf{Codex A}} & \rseventeenhead{\textbf{Codex B}} & \cellcolor{r17focus}\rseventeenhead{\textbf{LIFE}} \\
\midrule
\rowcolor{r17group}
\multicolumn{4}{l}{\rule{0pt}{12pt}\textcolor{r17ink}{\textbf{Question-level scores}}} \\
Two cups & $4/6$ & $5/6$ & \cellcolor{r17focus}$\mathbf{5.5/6}$ \\
Cushion handover and entry & $3/6$ & $5.5/6$ & \cellcolor{r17focus}$\mathbf{6/6}$ \\
Jacket location (asked at 12:30) & $\mathbf{6/6}$ & $\mathbf{6/6}$ & \cellcolor{r17focus}$\mathbf{6/6}$ \\
Early-morning visitor & $\mathbf{3/3}$ & $\mathbf{3/3}$ & \cellcolor{r17focus}$\mathbf{3/3}$ \\
First identified arrival & $2/3$ & $\mathbf{3/3}$ & \cellcolor{r17focus}$\mathbf{3/3}$ \\
End of the lunch break & $\mathbf{6/6}$ & $\mathbf{6/6}$ & \cellcolor{r17focus}$\mathbf{6/6}$ \\
\addlinespace[3pt]
\rowcolor{r17group}
\multicolumn{4}{l}{\rule{0pt}{12pt}\textcolor{r17ink}{\textbf{Overall performance}}} \\
Total score & $24/30$ & $28.5/30$ & \cellcolor{r17focus}$\mathbf{29.5/30}$ \\
Score (\%) & 80.0 & 95.0 & \cellcolor{r17focus}\textbf{98.3} \\
\addlinespace[3pt]
\rowcolor{r17group}
\multicolumn{4}{l}{\rule{0pt}{12pt}\textcolor{r17ink}{\textbf{Answer-time cost}}} \\
Answer time (s / question record) & 961.5 & 554.7 & \cellcolor{r17focus}\textbf{80.6} \\
Input tokens / record & 4.39M & 3.06M & \cellcolor{r17focus}\textbf{172.8K} \\
Cached input tokens / record & 4.26M & 2.91M & \cellcolor{r17focus}95.7K \\
Output tokens / record & 5,872.1 & 4,841.9 & \cellcolor{r17focus}\textbf{487.7} \\
\bottomrule
\end{tabular}
\endgroup
\end{table}

Across three runs per question, LIFE receives 29.5 of 30 possible points, Codex B receives 28.5, and Codex A receives 24. On several identity questions, Codex A located the relevant person or activity but did not provide a name by comparing against registration photos; Codex B used names available in the memory text. Codex B lost points in two runs, both on object reasoning: in one it attributed the cup of unestablished ownership to an attendee, and in the other it did not confirm that the cushion carried into the room was the user's. The scores therefore reflect the information available to each agent and its handling of identity.

Averaged over its 18 runs, LIFE takes 80.6 seconds per answer. Codex A takes 961.5 seconds and Codex B 554.7 seconds; the corresponding mean input totals are 172.8 thousand, 4.39 million, and 3.06 million tokens per answer. The A-to-B comparison shows how making the constructed memory available changes the coding agent's access to the recorded experience. These figures cover answering only and exclude memory construction; input totals include cached tokens.

Together, the cases show how LIFE connects continuous observation to later problem solving. RoboBeats selects moments for memory writing; Plastic Memory organizes descriptions and their supporting observations; and the execution agent combines memory access with available tools to answer questions about people, objects, places, and activities. The same evidence can be revisited and processed through different access methods as new questions arise, providing shared experience on which the capability-construction process of Section~\ref{sec:life-edm} can operate.

\section{Discussion}

LIFE connects continuous perception, long-term memory, and capability development. Our experiments examine proactive perception, memory question answering, and self-evolution, while the daily-life cases show how a robot can use observations from different times to answer questions about people, objects, and activities.

Moving from these capabilities to long-term autonomous operation requires further evaluation. As the environment changes, a robot must update its memory, resolve conflicts between old and new information, and use accumulated experience in later actions. Whether perception, memory, and capability development can work together reliably over long periods remains an open question.

Evaluating lifelong embodied intelligence also poses a challenge. Our real-robot study follows one robot through one continuous day. A day in the real world cannot be re-enacted, so comparing configurations or repeating runs on the same experience is possible only by recording the robot's stream and replaying it; replay, however, does not test how the robot's actions change what it later observes. Simulation platforms support controlled evaluation of embodied agents~\citep{savva2019habitat,tian2026abot}, but sustained environmental changes, natural human interaction, and the long-term consequences of actions remain difficult to capture fully. Multi-day online deployments with questions posed in situ, together with dynamic simulation, may help assess how robots accumulate and use experience in open environments.

\section{Conclusion and Future Work}

We presented LIFE, an embodied intelligence framework that connects proactive perception, long-term memory, and self-evolution. RoboBeats continuously monitors the environment and selects observations for memory writing; Plastic Memory retains experiences that can be revisited and reinterpreted; EDM turns methods developed during problem solving into reusable tools and skills. Under matched protocols, the gate improves condition tracking over its base model and, at the same writing budget, retains more useful evidence for later questions than uniform sampling. In ReSight's experiments, capabilities built with EDM carry over to held-out episodes and to new robot memories without further changes. At the scale of our real-robot case study, which replays one day of robot video with six questions, the integrated system answers everyday questions about people, objects, and activities more accurately and much faster than a coding agent working from the raw recording.

Future work will bring continuous perception, memory, long-term goals, and autonomous action closer together. A robot could organize its activities around an ongoing mission, choose what to attend to and where to explore, and adjust its actions in light of past experience. It could remember places it has visited, people it has met, and events it has observed, learning through continued interaction with the world.

Our goal is for embodied agents eventually to live autonomously in the real world, learning from their own experiences and building knowledge and capabilities over time. This is the direction we call \emph{Learning to Live}.

\bibliography{refs}
\clearpage
\appendix

\section{Contributors}
\label{app:contributors}

\newcommand{\contribrole}[1]{\textbf{\textcolor{agiblue}{#1}}}

\noindent\small
\setlength{\parindent}{0pt}\setlength{\parskip}{0.55em}

\contribrole{RoboBeats Training:} Lirong Che, Peidong Wang, Yimeng Su

\contribrole{Data Construction and Evaluation:} Peidong Wang, Lirong Che, Jiong Wang

\contribrole{Memory System:} Lirong Che, Yimeng Su

\contribrole{Tools \& Real-world Development:} Lirong Che, Jiong Wang, Peidong Wang, Yang Chen, Wenbo Fu, Zhenyu Huang, Tiancheng Xu

\vspace{1em}
\contribrole{Project Leaders}

Jian Su, Daqi Liu

\vspace{0.5em}
\contribrole{Supervisors}

Xueqian Wang\footnote{Tsinghua University.}, Yuzhe Yang, Xu Cao, Dongli Xie, Chuang Wang
\normalsize

\section{Proactive Perception}
\label{app:active-perception}

\subsection{Online Inference and Trigger States}
\label{app:robobeats-online}

\paragraph{Online processing.}
By default, the system receives frames at 1\,Hz. The model first reads the monitoring conditions, then processes timestamped images and their decision queries in order. It outputs a score for each condition at every step.

To avoid repeated computation, the model caches attention key and value states and recurrent states from the processed history. When a new frame arrives, the system appends the image and decision queries to the cached history. It computes the current predictions, updates the cache, and waits for the next frame.

Long streams are processed in chunks to bound the context length. Each chunk contains at most 100 frames, with 10 overlapping frames between adjacent chunks. At a chunk transition, the system rebuilds the cache from the condition descriptions and the last 10 frames of the preceding chunk, then continues with new frames. The overlapping frames retain their timestamps and provide preceding context. Only predictions for newly arriving frames are recorded, preserving a continuous sequence of outputs.

\paragraph{Trigger state update.}

\label{app:trigger-state}

For each condition $c_k$, the gate updates its binary state $s_{t,k}$ from score $p_{t,k}$ using separate activation and deactivation thresholds:
\begin{equation}
s_{t,k}=\begin{cases}
1, & p_{t,k}>0.5,\\
0, & p_{t,k}<0.3,\\
s_{t-1,k}, & \text{otherwise},
\end{cases}
\qquad s_{0,k}=0.
\label{eq:robobeats-hysteresis}
\end{equation}

\subsection{Evaluation Datasets and Held-out Results}
\label{app:robobeats-eval-datasets}

\paragraph{Dataset composition.}
For condition tracking, we use a held-out condition judgment set and public benchmarks. The held-out set contains 3,682 video windows excluded from training, with 70,027 supervised frames from eight data sources. It covers fixed-camera, third-person, and first-person views. Each window retains the interval in which a condition holds and the continuous observations before and after it. This tests generalization to held-out videos from sources covered by the training data. Public benchmarks cover temporal localization, condition triggering, and proactive responses. We evaluate online activity localization on 3,720 Charades-STA~\citep{gao2017tall} queries and first-trigger timing on 250 questions from the StreamingBench~\citep{lin2024streamingbench} proactive output task. The SPOT-Bench~\citep{chatterjee2026spot} action boundary detection task contains 2,340 response points across 322 videos and tests responses to the starts and ends of action steps. ESTP-Bench~\citep{zhang2025estp} uses all 2,264 questions over 14 tasks (1,212 single-question examples and 1,052 contextual questions) to evaluate proactive responses from a condition gate coupled to an answering model.

For event discovery, SSS-Bench tests whether observation selection supports subsequent memory question answering. The system selects and retains information without knowing future questions, then answers using the memory available at question time. We also evaluate retrospective question answering on 631 OVO-Bench~\citep{li2025ovobench} questions, comparing RoboBeats with the published OneStreamer results~\citep{zeng2026onestreamer}. The tasks cover episodic memory (EPM), action sequence identification (ASI), and hallucination detection (HLD).

\paragraph{Held-out condition judgment.}

\label{app:condition-heldout}

Table~\ref{tab:condition-heldout} reports overall and per-source results on held-out videos. Recall uses a threshold selected for a 10\% false-positive rate on the validation set.
\definecolor{ptsGroup}{HTML}{F1F3F5}
\definecolor{ptsInk}{HTML}{3C4650}
\definecolor{group0}{HTML}{F1F6FC}
\definecolor{ink0}{HTML}{173E5B}
\definecolor{group1}{HTML}{F1F8F4}
\definecolor{ink1}{HTML}{234832}
\definecolor{group2}{HTML}{F3F5F8}
\definecolor{ink2}{HTML}{354253}
\definecolor{group3}{HTML}{EFF8FA}
\definecolor{ink3}{HTML}{1C4D59}
\definecolor{focus3}{HTML}{D0EAF0}
\begin{table}[tp]
\centering
\caption{\textbf{Overall and per-source results on the held-out condition judgment set.}}
\label{tab:condition-heldout}
\setlength{\tabcolsep}{3pt}\renewcommand{\arraystretch}{1.12}\footnotesize
\resizebox{\linewidth}{!}{%
\begin{tabular}{l c *{11}{c}}
\toprule
\multirow{2}{*}{\textbf{Method}} & \multirow{2}{*}{\textbf{Size}} & \multicolumn{8}{c}{\textbf{Per-source AUROC $\uparrow$}} & \multicolumn{3}{c}{\textbf{Overall $\uparrow$}} \\
\cmidrule(lr){3-10}\cmidrule(lr){11-13}
 & & \textbf{Charades} & \textbf{TSU} & \textbf{Edinburgh} & \textbf{IKEA} & \textbf{\shortstack{Perception\\Test}} & \textbf{Ego4D} & \textbf{EGTEA} & \textbf{EPIC} & \textbf{AUROC} & \textbf{AP} & \textbf{Recall} \\
\midrule
\rowcolor{group0}\multicolumn{13}{l}{\textcolor{ink0}{\textbf{General VLMs}}} \\
\rowcolor{group0}Qwen3-VL-4B & 4B & 0.726 & 0.659 & 0.847 & 0.447 & 0.706 & 0.720 & 0.784 & 0.705 & 0.683 & 0.738 & 37.1\% \\
\rowcolor{group0}Qwen3-VL-8B & 8B & 0.727 & 0.686 & 0.849 & 0.404 & 0.709 & 0.711 & 0.811 & 0.702 & 0.680 & 0.739 & 39.3\% \\
\rowcolor{group0}InternVL3.5-8B & 8B & 0.728 & 0.668 & 0.799 & 0.443 & 0.724 & 0.731 & 0.812 & 0.728 & 0.670 & 0.730 & 36.2\% \\
\rowcolor{group0}MiniCPM-V-4.5 & 8B & 0.727 & 0.686 & 0.876 & 0.407 & 0.697 & 0.712 & 0.780 & 0.690 & 0.683 & 0.732 & 34.0\% \\
\rowcolor{group0}Qwen3.8-27B & 27B & 0.741 & 0.673 & 0.857 & 0.394 & 0.726 & 0.722 & 0.799 & 0.724 & 0.698 & 0.747 & 41.0\% \\
\addlinespace[3pt]
\rowcolor{group1}\multicolumn{13}{l}{\textcolor{ink1}{\textbf{Streaming Models}}} \\
\rowcolor{group1}MMDuet & 7B & 0.600 & 0.711 & 0.581 & 0.586 & 0.548 & 0.627 & 0.736 & 0.674 & 0.595 & 0.596 & 16.7\% \\
\rowcolor{group1}MMDuet2 & 3B & 0.670 & 0.545 & 0.582 & 0.491 & 0.647 & 0.664 & 0.682 & 0.662 & 0.634 & 0.640 & 22.8\% \\
\rowcolor{group1}Aha & 7B & 0.627 & 0.771 & 0.545 & 0.576 & 0.566 & 0.697 & 0.765 & 0.708 & 0.627 & 0.640 & 20.3\% \\
\rowcolor{group1}Em-Garde & 2B & 0.654 & 0.530 & 0.886 & 0.443 & 0.627 & 0.638 & 0.773 & 0.683 & 0.620 & 0.626 & 14.9\% \\
\addlinespace[3pt]
\rowcolor{group2}\multicolumn{13}{l}{\textcolor{ink2}{\textbf{Controls under Our Protocol}}} \\
\rowcolor{group2}Qwen3.5-0.8B & 0.8B & 0.469 & 0.577 & 0.492 & 0.525 & 0.492 & 0.494 & 0.542 & 0.528 & 0.487 & 0.508 & 9.6\% \\
\rowcolor{group2}Qwen3.5-2B & 2B & 0.565 & 0.633 & 0.698 & 0.479 & 0.600 & 0.584 & 0.506 & 0.572 & 0.612 & 0.602 & 14.1\% \\
\rowcolor{group2}Qwen3.5-4B & 4B & 0.603 & 0.591 & 0.611 & 0.488 & 0.602 & 0.602 & 0.642 & 0.606 & 0.616 & 0.616 & 20.8\% \\
\rowcolor{group2}Qwen3.8-27B & 27B & 0.652 & 0.495 & 0.833 & 0.564 & 0.620 & 0.627 & 0.669 & 0.691 & 0.642 & 0.648 & 22.2\% \\
\addlinespace[3pt]
\rowcolor{focus3}RoboBeats-0.8B & 0.8B & 0.840 & 0.946 & 0.918 & 0.876 & 0.881 & 0.863 & 0.917 & 0.848 & 0.897 & 0.904 & 68.0\% \\
\rowcolor{focus3}RoboBeats-2B & 2B & 0.853 & 0.942 & \textbf{0.948} & 0.905 & 0.903 & 0.893 & 0.918 & 0.885 & 0.915 & 0.923 & 73.6\% \\
\rowcolor{focus3}RoboBeats-4B & 4B & \textbf{0.858} & \textbf{0.948} & 0.936 & \textbf{0.929} & \textbf{0.906} & \textbf{0.902} & \textbf{0.923} & \textbf{0.900} & \textbf{0.925} & \textbf{0.930} & \textbf{76.2\%} \\
\bottomrule
\end{tabular}%
}

\end{table}

\subsection{Full Condition-Tracking Results}
\label{app:condition-full}

Tables~\ref{tab:charades-online-results}, \ref{tab:streaming-po-results}, and \ref{tab:spot} report the complete results for Charades-STA, StreamingBench proactive output, and SPOT-Bench, respectively, including the untuned base models and additional baselines.

\paragraph{Main comparison protocol.}
In Table~\ref{tab:condition-public-main}, published streaming methods use their original protocols. General VLMs and RoboBeats use our condition-judgment protocol on all three benchmarks: the model judges whether a condition holds at each step, and a fixed readout converts these judgments into outputs. The input rate is 1\,fps; SPOT-Bench uses start and end conditions at 2\,fps (Appendix~\ref{app:spot-readout}). RoboBeats training includes Charades-STA training data.

\begin{table}[tp]
\centering
\caption{\textbf{Charades-STA temporal localization results (\%).} Offline methods are published results that use the complete video. All online rows use our condition-judgment protocol at 1\,fps: from causal observations, the model judges at each second whether the query holds, and the same fixed readout converts these judgments into an interval. General VLMs see the most recent eight seconds. RoboBeats training includes Charades-STA training data. Bold marks the best result among our online evaluations.}
\label{tab:charades-online-results}
\footnotesize
\setlength{\tabcolsep}{3pt}
\renewcommand{\arraystretch}{1.06}
\definecolor{rbCharGroup}{HTML}{F1F3F5}
\definecolor{rbCharFocus}{HTML}{E3EDF8}
\begin{tabular}{llrrrr}
\toprule
 \textbf{Method} & \textbf{Readout} & \textbf{R@0.3} & \textbf{R@0.5} & \textbf{R@0.7} & \textbf{mIoU} \\
\midrule
\rowcolor{rbCharGroup}
\multicolumn{6}{l}{\textbf{Offline methods, published results}} \\
2D-TAN~\citep{zhang2019tan} & Offline & 57.3 & 45.8 & 27.9 & 41.0 \\
MMN~\citep{wang2021mmn} & Offline & 65.4 & 53.3 & 31.5 & 46.5 \\
TimeSuite~\citep{zeng2024timesuite} & Offline & 69.9 & 48.7 & 24.0 & -- \\
FV-Action (Qwen2.5-VL-7B)~\citep{huang2026fvaction} & Offline & 70.1 & 56.8 & 29.7 & -- \\
UniVTG~\citep{lin2023univtg} & Offline & 72.6 & 60.2 & 38.6 & 52.2 \\
Time-R1~\citep{wang2025timer1} & Offline & 78.1 & 60.8 & 35.3 & 58.1 \\
OmniVTG~\citep{zheng2026omnivtg} & Offline & 78.3 & 63.2 & 37.0 & -- \\
\midrule
\rowcolor{rbCharGroup}
\multicolumn{6}{l}{\textbf{General VLMs, condition-judgment protocol}} \\
Qwen3-VL-4B & Online & 40.6 & 22.1 & 9.2 & 26.3 \\
Qwen3.8-27B & Online & 43.0 & 24.0 & 10.3 & 28.2 \\
Qwen2.5-VL-7B & Online & 47.1 & 26.6 & 10.3 & 29.7 \\
Qwen3-VL-8B & Online & 46.6 & 27.3 & 11.2 & 30.2 \\
MiniCPM-V-4.5 & Online & 51.8 & 28.9 & 12.5 & 33.2 \\
InternVL3.5-8B & Online & 55.1 & 31.0 & 14.2 & 35.3 \\
\midrule
\rowcolor{rbCharGroup}
\multicolumn{6}{l}{\textbf{Base models, condition-judgment protocol}} \\
Qwen3.5-0.8B & Online & 33.2 & 0.5 & 0.2 & 25.5 \\
Qwen3.5-2B & Online & 32.1 & 7.4 & 2.9 & 22.6 \\
Qwen3.5-4B & Online & 29.8 & 9.9 & 4.6 & 20.8 \\
\midrule
\rowcolor{rbCharFocus}
RoboBeats-0.8B & Online & 72.7 & 53.1 & 26.9 & 47.8 \\
\rowcolor{rbCharFocus}
RoboBeats-2B & Online & 74.4 & \textbf{57.3} & 28.9 & 49.7 \\
\rowcolor{rbCharFocus}
RoboBeats-4B & Online & \textbf{74.7} & 57.2 & \textbf{30.1} & \textbf{50.3} \\
\rowcolor{rbCharFocus}
RoboBeats-0.8B & Offline & 78.4 & 59.5 & 31.7 & 52.0 \\
\rowcolor{rbCharFocus}
RoboBeats-2B & Offline & 79.6 & 62.7 & 33.3 & 53.9 \\
\rowcolor{rbCharFocus}
RoboBeats-4B & Offline & 81.3 & 64.2 & 34.5 & 55.0 \\
\bottomrule
\end{tabular}
\end{table}

\begin{table}[tp]
\centering
\caption{\textbf{StreamingBench proactive output (\%).} Proactive output reports first-trigger accuracy within the indicated time window. Published results use their original protocols. General VLMs, base models, and RoboBeats use our condition-judgment protocol at 1\,fps: the model judges at each second whether the condition holds, and the first trigger after the instruction is the response. General VLMs see an eight-second causal window. Bold marks the best result among our evaluations.}
\label{tab:streaming-po-results}
\footnotesize
\setlength{\tabcolsep}{2.2pt}
\renewcommand{\arraystretch}{1.12}
\definecolor{sbGroup}{HTML}{F1F3F5}
\definecolor{sbFocus}{HTML}{E3EDF8}
\begin{minipage}[t]{0.485\linewidth}
\vspace{0pt}
\begin{tabularx}{\linewidth}{>{\raggedright\arraybackslash}X rrr}
\toprule
 \textbf{Method} & \textbf{$\pm2$s} & \textbf{$\pm1$s} & \textbf{$\pm4$s} \\
\midrule
\rowcolor{sbGroup}\multicolumn{4}{l}{\textbf{Published, original protocols: closed-source}} \\
Claude Fable 5 & 71.6 & -- & -- \\
GPT-6 & 73.2 & -- & -- \\
Gemini 3.7 Flash & 80.0 & -- & -- \\
\rowcolor{sbGroup}\multicolumn{4}{l}{\textbf{Published, original protocols: open-source}} \\
Qwen3.5-4B (ProAgent base) & 11.6 & -- & -- \\
LiveStar-8B & 13.6 & -- & -- \\
MiniCPM-V-4.6 & 19.6 & -- & -- \\
Dispider & 25.3 & -- & -- \\
LLaVA-OneVision-7B & 28.8 & -- & -- \\
Qwen3-VL-8B-Instruct & 49.2 & -- & -- \\
ProAgent (StreamingJev-4B) & 57.2 & -- & -- \\
\bottomrule
\end{tabularx}
\end{minipage}\hfill
\begin{minipage}[t]{0.485\linewidth}
\vspace{0pt}
\begin{tabularx}{\linewidth}{>{\raggedright\arraybackslash}X rrr}
\toprule
 \textbf{Method} & \textbf{$\pm2$s} & \textbf{$\pm1$s} & \textbf{$\pm4$s} \\
\midrule
\rowcolor{sbGroup}\multicolumn{4}{l}{\textbf{General VLMs, condition-judgment protocol}} \\
InternVL3.5-8B & 39.2 & 26.4 & 48.8 \\
Qwen3-VL-4B & 39.6 & 30.0 & 51.2 \\
Qwen3-VL-8B & 44.8 & 30.4 & 54.8 \\
Qwen2.5-VL-7B & 45.6 & 33.2 & 54.8 \\
Qwen3.8-27B & 50.0 & 38.0 & 58.4 \\
MiniCPM-V-4.5 & 52.8 & 38.0 & 66.8 \\
\rowcolor{sbGroup}\multicolumn{4}{l}{\textbf{Base models, condition-judgment protocol}} \\
Qwen3.5-0.8B & 9.6 & 3.6 & 38.4 \\
Qwen3.5-2B & 22.8 & 14.0 & 47.2 \\
Qwen3.5-4B & 21.2 & 8.0 & 44.4 \\
\addlinespace[3pt]
\rowcolor{sbFocus}RoboBeats-0.8B & 51.2 & 40.0 & 65.6 \\
\rowcolor{sbFocus}RoboBeats-2B & 56.0 & 45.6 & 70.8 \\
\rowcolor{sbFocus}RoboBeats-4B & \textbf{64.0} & \textbf{53.6} & \textbf{78.0} \\
\bottomrule
\end{tabularx}
\end{minipage}
\end{table}

\definecolor{group0}{HTML}{F1F6FC}\definecolor{ink0}{HTML}{173E5B}
\definecolor{group1}{HTML}{F1F8F4}\definecolor{ink1}{HTML}{234832}
\definecolor{group2}{HTML}{F3F5F8}\definecolor{ink2}{HTML}{354253}
\definecolor{group3}{HTML}{D0EAF0}\definecolor{ink3}{HTML}{1C4D59}
\begin{table}[tp]\centering
\caption{\textbf{SPOT-Bench action boundary detection results.} All results use the official SPOT harness and scorer with strictly causal input. For general VLMs, base models, and RoboBeats, each request is rewritten as start and end conditions, the model judges them at 2\,fps (the input rate of the benchmark's streaming baselines), and a fixed readout emits the responses. Bold marks the best result among the models we evaluate.}\label{tab:spot}
\setlength{\tabcolsep}{3pt}\renewcommand{\arraystretch}{1.12}\footnotesize
\begin{tabular}{p{0.30\linewidth}*{5}{>{\centering\arraybackslash}p{\dimexpr(0.70\linewidth-12\tabcolsep)/5\relax}}}
\toprule
\textbf{Method} & \textbf{Size} & \textbf{T-score@5} & \textbf{T-F1@5} & \shortstack{\textbf{Ego-Exo4D}\\\textbf{T-F1@5}} & \shortstack{\textbf{HTStep}\\\textbf{T-F1@5}} \\
\midrule
\rowcolor{group0}\multicolumn{6}{l}{\textcolor{ink0}{\textbf{Published Results}}} \\
\rowcolor{group0} StreamingVLM & 7B & 4.8 & 3.9 & --- & --- \\
\rowcolor{group0} Qwen2.5-VL & 7B & 41.8 & 10.0 & --- & --- \\
\rowcolor{group0} Qwen3-VL & 8B & 40.5 & 10.7 & --- & --- \\
\rowcolor{group0} MMDuet2 & 3B & 13.9 & 14.2 & --- & --- \\
\rowcolor{group0} MiniCPM-o 4.5 & 9B & --- & 15.4 & --- & --- \\
\rowcolor{group0} JoyAI-VL-Interaction & 8B + 4B & 15.3 & 17.0 & --- & --- \\
\rowcolor{group0} Ctrl-SPOT & --- & --- & 22.2 & --- & --- \\
\addlinespace[3pt]
\rowcolor{group1}\multicolumn{6}{l}{\textcolor{ink1}{\textbf{General VLMs, condition-judgment protocol}}} \\
\rowcolor{group1} InternVL3.5-8B & 8B & 31.2 & 33.3 & 36.7 & 25.9 \\
\rowcolor{group1} MiniCPM-V-4.5 & 8B & 36.2 & 37.7 & 40.9 & 30.8 \\
\rowcolor{group1} Qwen2.5-VL-7B & 7B & 40.3 & 42.3 & 47.5 & 30.7 \\
\rowcolor{group1} Qwen3-VL-4B & 4B & 37.9 & 41.7 & 46.5 & 30.4 \\
\rowcolor{group1} Qwen3-VL-8B & 8B & 39.7 & 42.5 & 47.1 & 32.0 \\
\rowcolor{group1} Qwen3.8-27B & 27B & 40.7 & 43.0 & 47.4 & 33.3 \\
\addlinespace[3pt]
\rowcolor{group2}\multicolumn{6}{l}{\textcolor{ink2}{\textbf{Base Models, condition-judgment protocol}}} \\
\rowcolor{group2} Qwen3.5-0.8B & 0.8B & 2.5 & 4.5 & 5.9 & 1.0 \\
\rowcolor{group2} Qwen3.5-2B & 2B & 25.4 & 27.5 & 31.9 & 17.4 \\
\rowcolor{group2} Qwen3.5-4B & 4B & 25.7 & 26.4 & 32.3 & 13.7 \\
\addlinespace[3pt]
\rowcolor{group3} RoboBeats-0.8B & 0.8B & 39.7 & 40.7 & 45.1 & 31.0 \\
\rowcolor{group3} RoboBeats-2B & 2B & 42.9 & 43.8 & 48.5 & 33.5 \\
\rowcolor{group3} RoboBeats-4B & 4B & \textbf{44.8} & \textbf{45.8} & \textbf{51.3} & \textbf{33.6} \\
\bottomrule
\end{tabular}
\end{table}

\subsection{SPOT-Bench Output Construction}
\label{app:spot-readout}

For action-boundary detection, we rewrite each request as start and end conditions. The model scores whether each condition holds at 2\,fps, the input rate used by the benchmark's streaming baselines; the other condition-tracking benchmarks use 1\,fps. Fixed state transitions turn those scores into responses. The readout uses activation and deactivation thresholds of 0.5 and 0.3, takes the first rising edge for a start and the first falling edge after activation for an end, and emits each response at most once per question. This output construction limits repeated responses.

Published methods instead decide when to respond to the original question directly. To examine the effect of the output construction, we also evaluate general VLMs with the same rewritten conditions and fixed readout. Qwen2.5-VL-7B scores 10.0 T-F1@5 in the published official setting and 42.3 with our construction. These results compare the complete model-and-protocol configurations. The matched-protocol comparison isolates the effect of adaptation more directly: the Qwen3.5-2B base scores 27.5 and RoboBeats-2B scores 43.8.

\subsection{SSS-Bench Construction and Additional Results}
\label{app:sss-additional}\subsubsection{SSS-Bench Construction and Verification}
\label{app:sss-construction-details}

As shown in Figure~\ref{fig:robobeats-data}(B), construction comprises video and event-pool preparation, task and evidence selection, question and answer generation, strong-model verification, human review, and shortcut probing. We select continuous public video clips with recognizable actions and use native annotations and Qwen3.8-27B~\citep{qwen2026qwen38} teacher events as clues. Humans define question types and rules. GPT-6.1-sol inspects the full observation range at 1\,fps to determine the task scope, question time, and evidence. Early questions and answers are generated with templates, rules, and assistance from Qwen3.8-27B. Later questions, reference answers, and scoring rubrics are constructed by GPT-6.1-sol from the frames. Questions about order, occurrence number, and counts are checked against the images. An unannotated event is not assumed to be absent.

GPT-6.1-sol verifies candidates independently. It first answers from the full 1\,fps observation range with reference answers and event labels hidden, and reports evidence times. After the answer is saved, the references and scoring rubrics are revealed to check facts, evidence, and wording. Disputed questions undergo further review. Humans inspect cases and revise rules, followed by additional model verification. Qwen3.8-27B then answers using either the question alone or the question with a single frame at question time to screen for language and static-image shortcuts. These probes diagnose difficulty and do not guarantee that all shortcuts are absent.

Human review also refines scoring requirements and removes duplicates. The kitchen example in Figure~\ref{fig:robobeats-data}(B) illustrates the relation between historical evidence and a later question time.

\subsubsection{Additional Results}

\begin{table}[tp]
\centering
\caption{\textbf{SSS-Bench question-answering results (\%).} Native methods answer with their own systems; all other rows use Qwen3.8-27B as the reader. Direct-frame baselines see the sampled frames with no memory and no review. In the last group, the reader receives the text memory and may request one review window of the observation history, from which at most 10 frames are returned. Bold marks the best result within each group.}
\label{tab:memory-qa-writer-comparison}
\begingroup
\small
\definecolor{mqaSlate}{HTML}{F3F5F8}
\definecolor{mqaSlateInk}{HTML}{52647B}
\definecolor{mqaBlue}{HTML}{F1F6FC}
\definecolor{mqaBlueStrong}{HTML}{D5E5F7}
\definecolor{mqaBlueInk}{HTML}{245E86}
\definecolor{mqaMint}{HTML}{F1F8F4}
\definecolor{mqaMintStrong}{HTML}{D7ECDD}
\definecolor{mqaMintInk}{HTML}{37694B}
\setlength{\tabcolsep}{2pt}
\renewcommand{\arraystretch}{1.24}
\renewcommand{\tabularxcolumn}[1]{m{#1}}
\newcommand{\mqahead}[1]{{\fontsize{7.6}{9}\selectfont #1}}
\newcommand{\mqard}[1]{{\fontsize{8}{10}\selectfont #1}}
\begin{tabularx}{\linewidth}{%
  >{\raggedright\arraybackslash}m{0.235\linewidth}%
  >{\centering\arraybackslash}m{0.115\linewidth}%
  >{\hsize=0.9\hsize\linewidth=\hsize\centering\arraybackslash}X%
  *{2}{>{\hsize=1.25\hsize\linewidth=\hsize\centering\arraybackslash}X}%
  >{\hsize=1.4\hsize\linewidth=\hsize\centering\arraybackslash}X%
  *{4}{>{\hsize=0.8\hsize\linewidth=\hsize\centering\arraybackslash}X}}
\toprule
\multirow{2}{*}{\textbf{Method}} & \multirow{2}{*}{\textbf{Reader}} & \multirow{2}{*}{\mqahead{\shortstack{\textbf{Overall}\\(\%) $\uparrow$}}} & \multicolumn{3}{c}{\textbf{Question type}} & \multicolumn{4}{c}{\textbf{Observation duration (min)}} \\
\cmidrule(lr){4-6}\cmidrule(l){7-10}
 & & & \textbf{\mqahead{\shortstack{Temporal\\localization\\(247)}}} & \textbf{\mqahead{\shortstack{Event order\\\& repetition\\(202)}}} & \textbf{\mqahead{\shortstack{Object/person\\changes\\(169)}}} & \textbf{\mqahead{\shortstack{$<10$\\(33)}}} & \textbf{\mqahead{\shortstack{10--15\\(154)}}} & \textbf{\mqahead{\shortstack{15--20\\(240)}}} & \textbf{\mqahead{\shortstack{$\geq20$\\(191)}}} \\
\midrule
\rowcolor{mqaSlate}\multicolumn{10}{l}{\rule{0pt}{13pt}\textcolor{mqaSlateInk}{\textbf{Native end-to-end answering}}} \\
\rowcolor{mqaSlate}JoyAI-VL-Interaction & \mqard{Native} & \textbf{24.1} & 7.3 & \textbf{27.7} & \textbf{44.4} & \textbf{42.4} & \textbf{24.0} & \textbf{22.5} & \textbf{23.0} \\
\rowcolor{mqaSlate}OneStreamer & \mqard{Native} & 21.8 & \textbf{9.3} & 19.3 & 43.2 & 36.4 & 19.5 & 21.7 & 21.5 \\
\addlinespace[4pt]
\rowcolor{mqaSlate}\multicolumn{10}{l}{\rule{0pt}{13pt}\textcolor{mqaSlateInk}{\textbf{Direct frames, no memory}}} \\
\rowcolor{mqaSlate}Recent context (32 s) & \mqard{Qwen3.8-27B} & 13.6 & 1.2 & 16.3 & 28.4 & 39.4 & 9.1 & 12.9 & 13.6 \\
\rowcolor{mqaSlate}Offline uniform (64 frames) & \mqard{Qwen3.8-27B} & 39.3 & 22.7 & 38.1 & 65.1 & 63.6 & 38.3 & 40.8 & 34.0 \\
\rowcolor{mqaSlate}Offline uniform (128 frames) & \mqard{Qwen3.8-27B} & \textbf{50.0} & \textbf{41.3} & \textbf{43.6} & \textbf{70.4} & \textbf{75.8} & \textbf{51.9} & \textbf{48.8} & \textbf{45.5} \\
\addlinespace[4pt]
\rowcolor{mqaBlue}\multicolumn{10}{l}{\rule{0pt}{13pt}\textcolor{mqaBlueInk}{\textbf{Text memory only}}} \\
\rowcolor{mqaBlue}Qwen3.8-27B & \mqard{Qwen3.8-27B} & 40.9 & 32.4 & 35.1 & 60.4 & 51.5 & 45.5 & 37.1 & 40.3 \\
\rowcolor{mqaBlue}Uniform sampling & \mqard{Qwen3.8-27B} & 46.5 & 42.1 & 39.9 & 60.7 & 57.6 & 52.8 & 42.2 & 44.9 \\
\rowcolor{mqaBlueStrong}RoboBeats-0.8B & \mqard{Qwen3.8-27B} & 56.3 & 56.7 & 46.5 & 67.5 & \textbf{66.7} & 53.2 & 55.0 & \textbf{58.6} \\
\rowcolor{mqaBlueStrong}RoboBeats-2B & \mqard{Qwen3.8-27B} & \textbf{60.5} & \textbf{62.8} & \textbf{51.5} & \textbf{68.0} & 60.6 & \textbf{62.3} & \textbf{61.7} & 57.6 \\
\rowcolor{mqaBlueStrong}RoboBeats-4B & \mqard{Qwen3.8-27B} & 57.3 & 59.9 & 47.5 & 65.1 & \textbf{66.7} & 59.1 & 58.3 & 52.9 \\
\addlinespace[4pt]
\rowcolor{mqaMint}\multicolumn{10}{l}{\rule{0pt}{13pt}\textcolor{mqaMintInk}{\textbf{Text memory + up to 10 historical frames}}} \\
\rowcolor{mqaMint}OneStreamer PHCM & \mqard{Qwen3.8-27B} & 46.3 & 41.3 & 36.1 & 65.7 & 57.6 & 49.4 & 46.2 & 41.9 \\
\rowcolor{mqaMint}JoyAI-VL-Interaction & \mqard{Qwen3.8-27B} & 39.3 & 32.8 & 27.7 & 62.7 & 60.6 & 37.7 & 37.9 & 38.7 \\
\rowcolor{mqaMint}MMDuet (matched frames) & \mqard{Qwen3.8-27B} & 51.8 & 47.8 & 43.6 & 67.5 & 66.7 & 56.5 & 45.4 & 53.4 \\
\rowcolor{mqaMint}Qwen3.8-27B & \mqard{Qwen3.8-27B} & 49.7 & 47.0 & 40.6 & 64.5 & 69.7 & 50.0 & 47.5 & 48.7 \\
\rowcolor{mqaMint}Uniform sampling & \mqard{Qwen3.8-27B} & 54.7 & 54.3 & 44.7 & 67.3 & 72.7 & 59.7 & 50.0 & 53.4 \\
\rowcolor{mqaMintStrong}RoboBeats-0.8B & \mqard{Qwen3.8-27B} & 62.9 & 63.2 & \textbf{54.0} & 73.4 & \textbf{75.8} & 63.0 & 60.0 & \textbf{64.4} \\
\rowcolor{mqaMintStrong}RoboBeats-2B & \mqard{Qwen3.8-27B} & \textbf{63.1} & \textbf{66.4} & 49.5 & \textbf{74.6} & \textbf{75.8} & \textbf{66.9} & 62.5 & 58.6 \\
\rowcolor{mqaMintStrong}RoboBeats-4B & \mqard{Qwen3.8-27B} & 62.1 & \textbf{66.4} & 51.0 & 69.2 & 72.7 & 64.9 & \textbf{62.9} & 57.1 \\
\bottomrule
\end{tabularx}
\endgroup
\end{table}

\subsection{Additional Experiments on OVO-Bench}
\label{app:ovo-additional-experiments}

\paragraph{Memory sources and readers.}
We compare all-frame writing, OneStreamer PHCM, and RoboBeats-gated writing on the OVO-Bench Backward Tracing tasks. These module ablations are evaluated locally; they are separate from the published OneStreamer benchmark results in Table~\ref{tab:ovo-backward-tracing}. We use Qwen3.8-27B and OneStreamer-4B as readers and evaluate three answering modes using text memory only, recent frames only, and text memory combined with recent frames. For all-frame and RoboBeats-gated memories, Qwen3.8-27B writes the text from all sampled candidate frames or the RoboBeats-selected frames, respectively; OneStreamer PHCM uses OneStreamer-4B to write memory. Table~\ref{tab:memory-source-grouped} reports task accuracies and BT Avg., the equally weighted mean of EPM, ASI, and HLD. The text-memory and writing-volume comparisons in the main text use EPM+ASI Avg., the equally weighted mean of the two memory tasks.

\begingroup
\begin{table}[tp]\centering
\definecolor{blue}{HTML}{F1F6FC}\definecolor{bluefocus}{HTML}{D5E5F7}\definecolor{blueink}{HTML}{245E86}
\definecolor{mint}{HTML}{F1F8F4}\definecolor{mintfocus}{HTML}{D7ECDD}\definecolor{mintink}{HTML}{37694B}
\definecolor{slate}{HTML}{F3F5F8}\definecolor{slateink}{HTML}{52647B}
\caption{\textbf{Memory QA comparisons on OVO-Bench.} All rows are locally evaluated module configurations. All-frame and RoboBeats-gated memories are written by Qwen3.8-27B; RoboBeats selects the frames in the gated setting. OneStreamer PHCM uses OneStreamer-4B to write memory. All scores are percentages. BT Avg. averages EPM, ASI, and HLD equally. Bold marks the best result among memory sources for each reader and answer-time setting.}\label{tab:memory-source-grouped}
\setlength{\tabcolsep}{4.5pt}\renewcommand{\arraystretch}{1.15}\small
\begin{tabular}{l l *{4}{c}}
\toprule
\textbf{Reader} & \textbf{Memory source} & \textbf{EPM} & \textbf{ASI} & \textbf{HLD} & \textbf{BT Avg. $\uparrow$} \\
\midrule
\rowcolor{blue}\multicolumn{6}{l}{\rule[-5.5pt]{0pt}{18pt}\textcolor{blueink}{\textbf{Text memory only}}} \\
\rowcolor{blue}  & All frames & \textbf{51.18} & 64.19 & 98.39 & 71.25 \\
\rowcolor{blue}  & OneStreamer PHCM & 44.78 & 59.46 & \textbf{99.46} & 67.90 \\
\rowcolor{blue} \multirow{-3}{*}{OneStreamer-4B} & \cellcolor{bluefocus}RoboBeats-gated & \cellcolor{bluefocus}48.15 & \cellcolor{bluefocus}\textbf{68.24} & \cellcolor{bluefocus}98.39 & \cellcolor{bluefocus}\textbf{71.59} \\
\rowcolor{blue}  & All frames & \textbf{53.20} & \textbf{66.89} & \textbf{88.71} & \textbf{69.60} \\
\rowcolor{blue}  & OneStreamer PHCM & 41.08 & 59.46 & 85.48 & 62.01 \\
\rowcolor{blue} \multirow{-3}{*}{Qwen3.8-27B} & \cellcolor{bluefocus}RoboBeats-gated & \cellcolor{bluefocus}46.80 & \cellcolor{bluefocus}66.22 & \cellcolor{bluefocus}87.10 & \cellcolor{bluefocus}66.70 \\
\addlinespace[4pt]
\rowcolor{slate}\multicolumn{6}{l}{\rule[-5.5pt]{0pt}{18pt}\textcolor{slateink}{\textbf{Recent frames only}}} \\
\rowcolor{slate} OneStreamer-4B &  & 62.29 & 64.19 & 99.46 & 75.31 \\
\rowcolor{slate} Qwen3.8-27B & \multirow{-2}{*}{No memory} & 64.98 & 63.51 & 73.66 & 67.38 \\
\addlinespace[4pt]
\rowcolor{mint}\multicolumn{6}{l}{\rule[-5.5pt]{0pt}{18pt}\textcolor{mintink}{\textbf{Text memory + recent frames}}} \\
\rowcolor{mint}  & All frames & \textbf{65.32} & 66.89 & \textbf{99.46} & 77.22 \\
\rowcolor{mint}  & OneStreamer PHCM & 61.28 & 69.59 & \textbf{99.46} & 76.78 \\
\rowcolor{mint} \multirow{-3}{*}{OneStreamer-4B} & \cellcolor{mintfocus}RoboBeats-gated & \cellcolor{mintfocus}63.97 & \cellcolor{mintfocus}\textbf{71.62} & \cellcolor{mintfocus}\textbf{99.46} & \cellcolor{mintfocus}\textbf{78.35} \\
\rowcolor{mint}  & All frames & \textbf{67.68} & \textbf{72.30} & \textbf{72.04} & \textbf{70.67} \\
\rowcolor{mint}  & OneStreamer PHCM & 65.32 & 68.24 & 69.89 & 67.82 \\
\rowcolor{mint} \multirow{-3}{*}{Qwen3.8-27B} & \cellcolor{mintfocus}RoboBeats-gated & \cellcolor{mintfocus}62.29 & \cellcolor{mintfocus}70.95 & \cellcolor{mintfocus}70.43 & \cellcolor{mintfocus}67.89 \\
\bottomrule
\end{tabular}\end{table}
\endgroup

\paragraph{Question answering performance and token usage with DeltaStreamer.}
We reproduce DeltaStreamer using Qwen3.8-27B for memory writing and question answering under the OVO-Bench evaluation protocol. We then integrate the RoboBeats event gate to select observations for the writer. Table~\ref{tab:deltastreamer-robobeats} compares question answering accuracy and token usage of Qwen3.8-27B before and after integration. Adding RoboBeats reduces BT Avg. by 3.19 percentage points, while reducing input and output tokens by 56.31\% and 49.17\%, respectively. Event-discovery gating therefore substantially reduces processing by the large model in the memory pipeline, with a modest decrease in question answering accuracy.

\begingroup
\begin{table}[tp]\centering
\definecolor{focus}{HTML}{E3EDF8}
\caption{\textbf{QA performance (\%) and token usage of DeltaStreamer.} BT Avg. averages EPM, ASI, and HLD equally. Qwen3.8-27B is used for memory writing and answering.}\label{tab:deltastreamer-robobeats}
\setlength{\tabcolsep}{5pt}\renewcommand{\arraystretch}{1.18}\small
\begin{tabular}{l *{6}{c}}
\toprule
\textbf{Method} & \textbf{BT Avg. $\uparrow$} & \textbf{EPM $\uparrow$} & \textbf{ASI $\uparrow$} & \textbf{HLD $\uparrow$} & \textbf{\shortstack{Input tokens\\(M) $\downarrow$}} & \textbf{\shortstack{Output tokens\\(M) $\downarrow$}} \\
\midrule
DeltaStreamer & \textbf{59.19} & 52.86 & \textbf{68.24} & \textbf{56.45} & 2,925.2 & 63.2 \\
\rowcolor{focus}DeltaStreamer + RoboBeats & 56.00 & \textbf{54.88} & 64.19 & 48.92 & \textbf{1,278.0} & \textbf{32.1} \\
\bottomrule
\end{tabular}\end{table}
\endgroup

\FloatBarrier

\section{Memory Interfaces and Implementation Details}
\label{app:life-memory}
\setcounter{equation}{0}
\renewcommand{\theequation}{\Alph{section}.\arabic{equation}}

This appendix expands the keyframe interface, view assignment, retrieval, and historical-access rules in Section~\ref{sec:life-memory-views}. ReSight provides the original fields, tools, and EDM details~\citep{che2026resight}.

\subsection{Keyframe Fields and Writing Rules}
\label{app:life-keyframes}

Every keyframe has an index record with the same fields, whether it was sampled along a route or saved by proactive perception. The source and trigger record how a frame was written; they do not determine which fields it has. Derived fields are produced by builders that run over every retained frame, so an empty field means that the corresponding builder has not yet processed the frame. Table~\ref{tab:life-keyframe-fields} lists the fields.

\begin{table}[tp]
\centering
\caption{\textbf{Fields in a keyframe record.} All keyframes share these fields. Source and trigger describe how a frame was written; derived fields come from builders applied to every frame.}
\label{tab:life-keyframe-fields}
\small\setlength{\tabcolsep}{5pt}
\renewcommand{\arraystretch}{1.18}
\begin{tabularx}{\linewidth}{lX}
\toprule
\rowcolor[HTML]{F1F3F5}\textbf{Field} & \textbf{Content} \\
\midrule
\texttt{ref} & Frame reference identifying the stored image. \\
\texttt{t} & Capture time. \\
\texttt{pose} & Capture position, body heading, head rotation, and map. When the writer does not supply a pose, the pose timeline is queried at the capture time. \\
\texttt{source}, \texttt{trigger} & How the frame was written (route sampling, on-site user confirmation, or proactive perception) and the reason, such as a state change, an event, state maintenance, a fired condition, or a person's arrival. \\
\texttt{event} & Activity associated with the write: type, identifier, and text, such as a state or event description, the original condition, or a navigation task. \\
\texttt{segment}, \texttt{tracks} & Perceptual state segment covering the capture time and the tracks of people present. \\
\texttt{fields} & Derived fields: scene description, object list, image text, and text and image embeddings; visible people and counts of unrecognized faces; detection boxes; recognized registered objects, including presence, holder, and owner; view and segment membership; and the versions of the builders that produced them. \\
\bottomrule
\end{tabularx}
\end{table}

A keyframe is written when it is expected to add information to memory (Equation~\eqref{eq:life-memory-write}), and different triggers estimate this gain in different ways. Along a route, the robot must have moved or turned far enough since the last accepted frame, the image must be sharp and not captured during a fast turn, and candidates whose viewing frusta overlap accepted frames are excluded; these checks estimate viewpoint gain. Proactive perception writes a frame when a state changes, an event occurs, a condition fires, or a person arrives, which estimates event gain. When a perception inference finishes without a state change, a state-maintenance frame records that the state still holds. Frames confirmed by a user on site are written directly.

Any keyframe can serve as a spatial frame of its view, representing what the place looks like, if it passes the same quality check regardless of how it was written: the image is sharp, the robot is neither turning nor moving, people occlude little of the scene, and the frame adds information relative to the view's existing spatial frames. A frame that passes becomes the latest spatial frame of its view. Frames that do not pass remain members of the view and serve as evidence of what happened there.

\subsection{View Assignment and Viewing Direction}
\label{app:life-view-geometry}

The assignment operator $\sigma$ in Equation~\eqref{eq:life-view-assignment} uses position, direction, and an estimated gaze region within a map. Let $p_i=(\mathbf{x}_i,\psi_i)$, where $\mathbf{x}_i$ is a planar position and $\psi_i$ is the viewing direction: body heading plus head rotation when available, approximating the planar direction of the camera. Angular differences use the minimum circular separation to avoid discontinuities across $\pm180^\circ$:
\begin{equation}
\begin{aligned}
\mathrm{same}(p_i,p_j)
&=\mathbf{1}\!\left[\lVert\mathbf{x}_i-\mathbf{x}_j\rVert<d\ \wedge\ d_{\mathrm{ang}}(\psi_i,\psi_j)<\delta\right]\\
&\quad\vee\ \mathbf{1}\!\left[\lVert\mathbf{a}_i-\mathbf{a}_j\rVert<\rho\right],\\
\mathbf{a}_i&=\mathbf{x}_i+g\,(\cos\psi_i,\sin\psi_i).
\end{aligned}
\label{eq:life-same-view}
\end{equation}
Here, $d_{\mathrm{ang}}$ denotes the minimum circular angle, and $\mathbf{a}_i$ is a reference point at distance $g$ along direction $\psi_i$, approximating the gaze region. This point is geometrically constructed and is not a detected object's position. The second criterion permits observations from different positions to share a view when their reference points are nearby. Among views satisfying the criteria, a new frame selects the anchor with the lowest position-and-direction matching cost; otherwise, it creates a new view. Frames from different maps are assigned separately.

\subsection{Retrieval Fusion and Appearance-Change Cues}
\label{app:life-view-retrieval}

The score $s(q,i)$ in Equation~\eqref{eq:life-view-retrieval} combines retrieval channels by rank fusion. Channels are defined by field type: keyword matching on image text, text embeddings of descriptions and object lists, and image embeddings. Descriptions written by different models are scored in separate channels because they do not share a score scale: English scene descriptions and the writer's Chinese state and event sentences score differently against the same English query. The image-embedding channel scores every keyframe on one scale. Queries containing rare terms, such as room numbers, prioritize frames containing all rare terms. Place queries rank views by their spatial frames: the highest-ranked spatial frame of each view becomes $r_q(v)$, accompanied by the latest observation $c(v)$. Other members of the view that match the query are returned with it as evidence of what happened there.

Appearance-change cues compare SigLIP2 image embeddings. The comparison in Equation~\eqref{eq:life-view-retrieval} is made only when both embeddings are available and the latest observation is clearly later than the matching frame. Otherwise, the pair is unassessed. The cue reflects overall appearance differences; identifying their cause requires source-image inspection.

\subsection{Representative-Frame Selection and Reclamation}
\label{app:life-retention}

The view-retention policy preserves the earliest and latest observations and greedily selects additional representatives during nightly maintenance. The newly selected set $S$ is limited to $B$ frames. Frames retained for other holders are exempt from this allowance, so $B$ is not a bound on the total number of frames in a view.
\begin{equation}
\begin{aligned}
F_v(S)&=\sum_{i\in U_v}w_i\max_{j\in S\cup H_v}k(i,j)
+\beta\log\det\!\left(I+\alpha K_{S\cup H_v}\right),\\
k(i,j)&=e^{-(1-\cos(\mathbf{z}_i,\mathbf{z}_j))/\tau_v}\,
         e^{-|\log a_i-\log a_j|/\tau_t}.
\end{aligned}
\label{eq:life-view-retention}
\end{equation}
$U_v$ contains candidate frames older than a minimum retention age. $H_v$ contains the earliest and latest observations and full frames required by holders such as event segments, episodes, or tasks. These frames contribute to coverage but do not count toward the allowance for $S$. $K$ is the similarity matrix induced by $k$, and $w_i$ defaults to one. Frame age $a_i$ is measured in hours and has a lower bound.

The coverage term favors frames that represent other observations, while the log-determinant term encourages diversity. The kernel combines visual similarity with differences in log age. For the same absolute time interval, recent observations have a larger log-age separation. This encourages finer retention of recent changes, although selection also depends on visual content, existing held frames, and stopping criteria. $\tau_v$ and $\tau_t$ set the visual and temporal similarity scales, $\alpha$ and $\beta$ weight the diversity term, and selection stops when the marginal gain falls below a threshold.

The latest frame of a view is determined by temporal ordering at query time, so records carry no mutable ``current'' flag; when a new frame joins a view, the previous latest frame becomes historical. Frames no longer referenced by any holder become reclamation candidates. Representative selection and physical deletion are separate steps: the collector first produces a plan that deletes or downgrades candidates, and storage is released when the plan is applied.

\subsection{Accessing a View's History}
\label{app:life-view-history}

The history interface specifies a view using the robot's current viewing direction, a navigation point, or the capture pose of a historical frame. Given requested times $T$, it seeks the nearest retained observation in that view for each time:
\begin{equation}
\mathcal{H}(v,T)=\left\{\arg\min_{i\in V_v^{\mathrm{keep}}}|t_i-u|:u\in T\right\}.
\label{eq:life-history}
\end{equation}
$V_v^{\mathrm{keep}}$ is the set of retained frames for the view. Equation~\eqref{eq:life-history} defines nearest-time matching over a nonempty candidate set. Returned evidence is tied to its actual observation time, not necessarily the exact requested time. When $T$ is empty, the interface lists the view's retained frames in chronological order.

If no suitable retained frame exists near a requested time, the tool can consult the pose timeline to check whether the robot was at the relevant observation location, then attempt to read a frame from the raw-video buffer. Recovery depends on the recording still being retained and accessible. A request for ``now'' can acquire a new frame when the robot is at the specified observation location.

Historical comparison uses detections from the first and last frames. When the capture poses nearly coincide, same-category boxes are matched by image position to indicate continued visibility, position differences, objects no longer detected, and newly detected objects. For larger pose differences, only category counts are compared. Registered objects additionally use their event ledger to list recorded appearances, movements, and departures during the interval. Targets outside the vocabulary can be detected by supplying a temporary English category. These differences require interpretation against source images so that occlusion or missed detection is not treated as evidence of disappearance.

\section{24-Hour Real-Robot Experiment Details}
\label{app:real-robot-details}

\paragraph{Memory construction.}
The recorded day runs from 21:00 on October 21 to 21:00 on October 22, 2026. The event-discovery gate and the background-difference prefilter select 256 windows for the initial Qwen3.8-27B writer, six of them during the night, and the resulting memory retains 1,296 keyframes. The consolidation process, which organizes the writer's output in half-hour blocks, uses 363 writer calls with approximately 8.04 million input and 0.14 million output tokens on the local Qwen3.8-27B model. Object detection and tracking and the condition-based rescans add further construction cost that these counts do not include.

\paragraph{Scoring and inputs.}
Every configuration answers each of the six questions three times, and each answer is manually scored against the recorded rubric, with partial credit for some answers. LIFE's execution agent receives the two turns of a two-turn question in sequence, whereas the coding agents receive both turns together as one numbered prompt. Table~\ref{tab:real-robot-comparison} reports per-question scores.
\FloatBarrier

\end{document}